\PassOptionsToPackage{table}{xcolor} 
\documentclass{article} % For LaTeX2e
\usepackage{iclr2026_conference,times}

\usepackage{amsmath,amsfonts,bm}

\def\eqref#1{equation~\ref{#1}}
\def\1{\bm{1}}

\DeclareMathAlphabet{\mathsfit}{\encodingdefault}{\sfdefault}{m}{sl}
\SetMathAlphabet{\mathsfit}{bold}{\encodingdefault}{\sfdefault}{bx}{n}

\usepackage{hyperref}
\usepackage{url}

\usepackage{textcomp}
\usepackage{stfloats}
\usepackage{url}
\usepackage{verbatim}
\usepackage{graphicx}
\usepackage{multirow}
\usepackage{multicol}
\usepackage{tabularx}
\usepackage{booktabs}
\usepackage[linesnumbered,ruled,vlined]{algorithm2e}
\usepackage{pifont}
\usepackage{chngcntr}
\title{Latent Noise Mask for Reducing Visual Redundancy in Multimodal Large Language Models}

\author{
Kai Jiang$^{1,2}$, Ruishu Zhu$^{1,2}$, Siqi Huang$^{3,2}$, Hongyuan Zhang$^{4,2,\dagger}$ \& Xuelong Li$^{2,\dagger}$\\
$^1$School of Artificial Intelligence, OPtics and ElectroNics (iOPEN),\\
Northwestern Polytechnical University\\
$^2$Institute of Artificial Intelligence, China Telecom (TeleAI)\\
$^3$Fudan University\\
$^4$The University of Hong Kong\\
$^\dagger$Corresponding authors
}

\iclrfinalcopy

\begin{document}

\maketitle

\begin{abstract}
Multimodal large language models (MLLMs) often fail in fine-grained visual reasoning, as question-relevant visual cues are diluted by dense and redundant image tokens. Recent multimodal reasoning methods usually extend chain-of-thought from language models into visual or latent spaces, seeking to add intermediate reasoning states while overlooking the negative impact of redundant visual tokens. We propose \textbf{L}at\textbf{E}nt \textbf{N}oise ma\textbf{S}k (\textsc{Lens}), a question-conditioned visual evidence purification framework that empowers MLLMs to reason with cleaner visual cues in latent space. \textsc{Lens} introduces a lightweight \emph{Lens Evidence Token} (LET) to score which visual tokens support the current question and preserve them during decoding. Guided by the LET scores, it injects adaptive latent noise into low-relevance tokens, softly suppressing distractors without changing the model backbone or token sequence. With only one temporary learnable control token and a lightweight noise generator, \textsc{Lens} adds minimal overhead while improving the base MLLM by 2.4--6.4 points on most VQA datasets and by 4.1--6.4 points on grounding tasks. These results show that multimodal reasoning can benefit more directly from cleaner question-relevant visual evidence than from simply extending the reasoning trace.
\end{abstract}

\section{Introduction}
Multimodal large language models (MLLMs) have become a central interface for visual understanding and multimodal reasoning~\citep{qwen3vl, llava-ov, internvl3_5}. By mapping images into token sequences that can be processed by large language models, they inherit strong language reasoning ability and support tasks such as visual question answering, spatial reasoning, chart understanding, and visual grounding. However, reliable visual reasoning \emph{often depends on a few local cues rather than the whole image}. Accordingly, many failures arise when small objects, fine-grained attributes, or spatial relations are diluted by background regions, nearby distractors, or language priors~\citep{wang2025perception, zhang2024mathverse}. \textbf{The key bottleneck is not only weak reasoning but also the difficulty of isolating question-relevant evidence from dense visual tokens.}

\begin{figure*}[!t]
    \centering
    \includegraphics[width=5.3in]{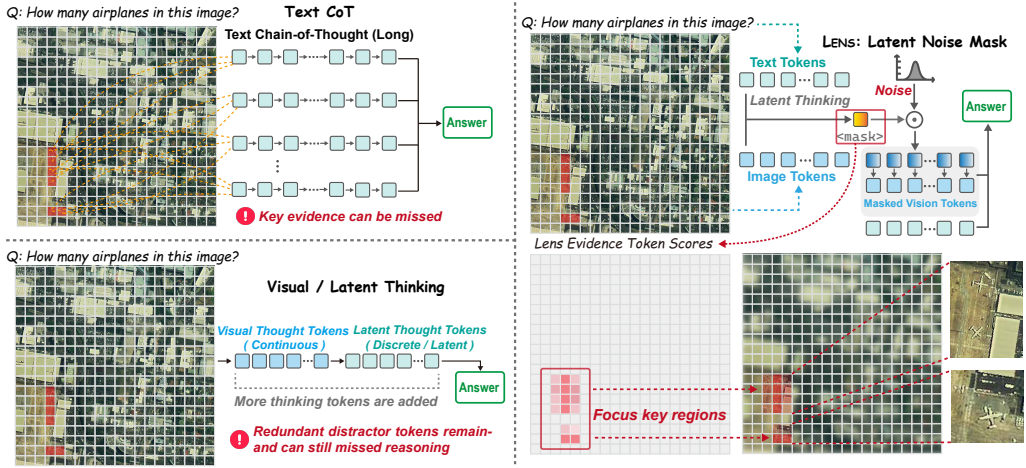}
\caption{Motivation of \textsc{Lens}. Existing multimodal reasoning methods often add visual or latent reasoning states, yet redundant image tokens can still distract the model. \textsc{Lens} learns \emph{Lens Evidence Token} (LET) to score visual evidence and softly suppress irrelevant visual tokens in latent space.} \label{fig:motivation}
%\vspace{-10pt}
\end{figure*}

Recent works address this gap by making MLLMs reason longer or reason in richer spaces. Textual chain-of-thought prompting encourages step-by-step inference and has been extended to multimodal tasks~\citep{wei2022chain, zhang2023multimodal, RN}. Other methods introduce visual cues through image crops, bounding boxes, visual tools, or image generation~\citep{shao2024visual, zheng2025deepeyes, su2025openthinkimg}. Latent-space methods further move intermediate reasoning into continuous representations, e.g., Mirage interleaves latent visual tokens with text, DMLR refines latent think tokens and injects relevant patches, and VisMem stores latent vision memories during generation~\citep{yang2025machine, liu2025reasoning, yu2025vismem}. \textbf{Most recent multimodal reasoning methods improve reasoning by adding more textual, visual, or latent context.} These studies show that visual and latent spaces can support reasoning beyond pure text, but they fail to directly reduce the influence of irrelevant image tokens.

However, this additive view leaves a simpler question underexplored. Images are already encoded as dense visual token sequences, and many tokens are unrelated to a specific question. For instance, when answering the color of a small object, background tokens and unrelated objects may dominate the context. When grounding a target object, visually similar neighbors may provide strong but misleading cues. \textbf{When the needed evidence is already present, the main challenge becomes visual selectivity rather than visual availability.} In such cases, adding more reasoning context not only leaves distractors untouched, but can also make decoding harder by increasing the amount of irrelevant context that the model must sift through.

In this paper, we study multimodal reasoning from the perspective of visual evidence purification. \textbf{\textsc{Lens} learns which visual tokens support the current question and suppresses the remaining tokens before they interfere with decoding.} This differs from token pruning~\citep{yu2025introducing, bigverdi2025perception, MuNG} or hard region selection~\citep{shao2024visual, fu2025refocus, ViewMask}. Removing tokens can be brittle when the relevance prediction is imperfect, and it may change the token layout expected by the backbone. Instead, we use a soft latent-space intervention that keeps the original sequence structure while reducing the influence of distractor tokens. As shown in Fig.~\ref{fig:motivation}, the goal is not to generate longer textual thoughts or add more visual thoughts, but to clean the visual evidence that the model already receives.

To this end, we propose Latent Noise Mask, denoted as \textsc{Lens}, a question-conditioned visual evidence suppression framework for MLLMs. \textbf{\textsc{Lens} provides a lightweight question-conditioned intervention that purifies visual evidence without external tools, costly rationales, or backbone modification.} It introduces a temporary \emph{Lens Evidence Token} (LET) that estimates the relevance of each visual token to the current question. Instead of using expensive chain-of-thought traces, \textsc{Lens} uses object-level annotations as evidence supervision by mapping question-relevant boxes to visual patches. Guided by the LET scores, it preserves evidence tokens and injects adaptive latent noise into low-relevance tokens. The backbone and token sequence structure remain unchanged, making the framework easy to attach to existing MLLMs.

\begin{figure*}[!t]
    \centering
    \includegraphics[width=5.3in]{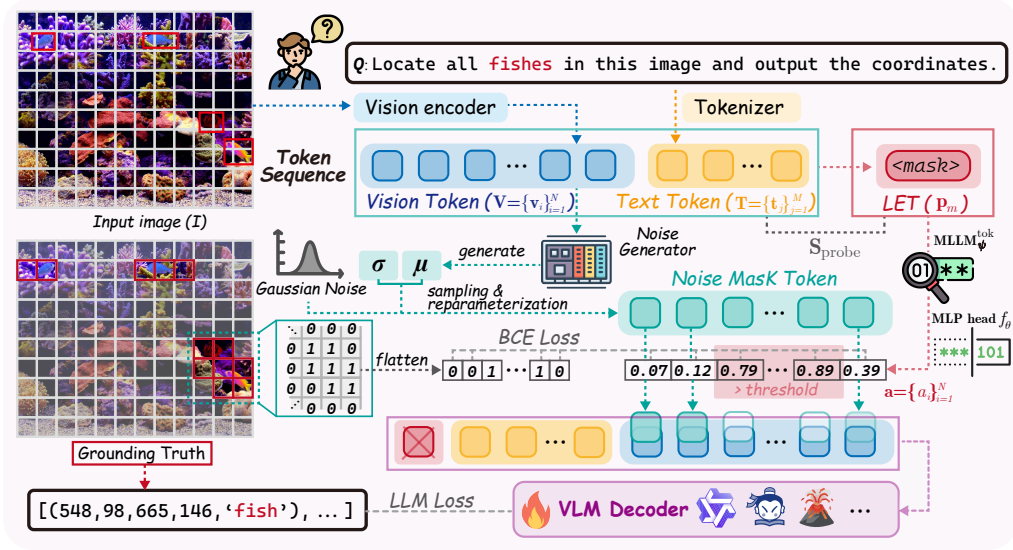}
    \caption{Overview of \textsc{Lens}. Evidence probing appends a temporary $\langle \mathrm{mask} \rangle$ token, whose hidden state is decoded by a simple MLP head $f_{\theta}$ into the LET scores. A threshold $\tau$ converts the scores into a suppression gate, and a small noise generator predicts token-specific latent noise for low-gate visual tokens. The control token is removed before final decoding, so the MLLM receives the purified sequence $\widetilde{\mathbf{V}}$ with the original token layout.} \label{fig:overview}
%\vspace{-10pt}
\end{figure*}

Our experiments show that this shift from visual addition to visual suppression is effective. Across VQA and grounding benchmarks, \textsc{Lens} improves the base MLLM by 2.4--6.4 points on most VQA datasets and by 4.1--6.4 points on grounding tasks. Qualitative results further show that the predicted LET scores align with question-relevant regions and reduce attention to distractors. \textbf{The results show that cleaner question-relevant visual evidence can improve multimodal reasoning more directly than simply extending the reasoning trace.} Our contributions are summarized as follows:
%\begin{itemize}
%    \item We identify a visual evidence bottleneck: dense image tokens contain the needed cues, but question-irrelevant distractors compete with them during decoding. We connect this bottleneck to grounding and ablation analyses that show the benefit of reducing distractor influence.
%    \item We propose \textsc{Lens}, a question-conditioned latent noise masking framework that uses a \emph{Lens Evidence Token} to learn token-level evidence scores from object-level supervision and softly suppresses irrelevant visual tokens without changing the MLLM backbone.
%    \item We demonstrate that \textsc{Lens} improves VQA and grounding performance while adding only a temporary learnable control token and a lightweight suppression module, with the backbone and token sequence kept unchanged.
%\end{itemize}
\begin{itemize}
\item We reveal that question irrelevant visual distractors in dense visual tokens can compete with key evidence and hinder fine grained visual reasoning, which motivates us to explore for suppressing visual interference than simply adding more tokens with visual information. 
\item We propose a lightweight visual evidence purification framework that learns question conditioned token relevance and softly suppresses low relevance visual tokens in latent space.
\item Experiments show the advantage in both accuracy and efficiency over other methods, demonstrating the effectiveness and generality of reasoning with cleaner visual evidence.
\end{itemize}

\section{Related Work}
\paragraph{Visual and multimodal reasoning}
Chain-of-thought prompting improves language reasoning by exposing intermediate rationales~\citep{wei2022chain, MiN}, and has been adapted to multimodal tasks that require visual evidence~\citep{zhang2023multimodal, mondal2024kam}. Recent MLLM methods add visual evidence through grounded rationales, selected regions, structured prompts, visual sketching, image editing, or tool use~\citep{shao2024visual, gao2025interleaved, mitra2024compositional, hu2024visual, zheng2025deepeyes, su2025openthinkimg, fu2025refocus, li2025imagine, zhao2025pyvision, wang2025pixel, fan2025grit}. These methods show that explicit visual context can improve perception and reasoning, but they usually expand the reasoning trace or modify the visual input. \textbf{\textsc{Lens} addresses a complementary problem. It keeps the original image sequence and suppresses question-irrelevant visual tokens before decoding.}

\paragraph{Latent perception and token-level intervention}
Latent-space reasoning moves intermediate computation into continuous states, reducing the cost of generating long textual chains~\citep{hao2024training, li2025seek, VPN}. Multimodal latent methods further interleave, optimize, predict, or retrieve latent visual context for reasoning~\citep{yang2025machine, liu2025reasoning, qin2025chain, yu2025vismem, rne}. Perception-aware and token-level methods also improve visual reasoning through reinforcement learning, perception objectives, selected visual tokens, or visual prompts~\citep{liu2025visual, shen2025vlm, huang2025vision, wang2025perception, chen2025mint, lei2025scaffolding, yu2025introducing, bigverdi2025perception}. \textsc{Lens} is closest to this line, but differs in the intervention target. Rather than adding memory, retrieving patches, or pruning tokens, \textbf{it introduces a question-conditioned \emph{Lens Evidence Token} and applies latent noise to tokens with low LET scores while preserving the original token layout.}

\section{Methodology}
\subsection{Problem Setup and Method Overview}
\textbf{Problem Setup.}\quad We consider an MLLM that receives an image--question pair $(I,Q)$. The image encoder maps $I$ into visual tokens $\mathbf{V}=\{\mathbf{v}_i\}_{i=1}^{N}$, and the tokenizer maps $Q$ into text tokens $\mathbf{T}=\{\mathbf{t}_j\}_{j=1}^{M}$. A standard MLLM decodes the answer sequence $Y=\{y_l\}_{l=1}^{L}$ from
\begin{align}
P_{\psi}(Y \mid \mathbf{V},\mathbf{T})
=
\prod_{l=1}^{L}
P_{\psi}(y_l \mid \mathbf{V},\mathbf{T},y_{<l}).
\end{align}
This formulation gives all visual tokens the same access to the decoder. In many visual reasoning tasks, however, only a small subset of tokens provides direct evidence for the question. The remaining tokens often describe background regions, unrelated objects, or similar distractors. These tokens can weaken the useful evidence before answer generation.

%\begin{align}
%P_{\psi}(Y \mid \widetilde{\mathbf{V}},\mathbf{X}).
%\end{align}
%The original visual token count and backbone interface are preserved.
%\textsc{Lens} aims to purify the visual context before decoding. It introduces a question-conditioned \emph{Lens Evidence Token} to produce an evidence score vector $\mathbf{a}=\{a_i\}_{i=1}^{N}$, where $a_i\in[0,1]$ measures the reliability of visual token $\mathbf{v}_i$ for the current question. Guided by these LET scores, \textsc{Lens} constructs a purified visual sequence $\widetilde{\mathbf{V}}=\{\widetilde{\mathbf{v}}_i\}_{i=1}^{N}$. 
%The final answer is then decoded as $P_{\psi}(Y \mid \widetilde{\mathbf{V}},\mathbf{T})$, while the original visual token count and backbone interface are preserved.
%
%\textbf{Method Overview.} As illustrated in Fig.~\ref{fig:motivation} and~\ref{fig:overview}, \textsc{Lens} has two coupled stages. \textbf{Evidence probing} appends a temporary $\langle mask\rangle$ control token $\mathbf{p}_m$ to the joint input $[\mathbf{V},\mathbf{T}]$; the hidden state of this token is decoded into the LET scores $\mathbf{a}\in[0,1]^N$, and the control token is removed afterwards. \textbf{Latent suppression} converts $\mathbf{a}$ into a suppression gate $g_i$ and forms $\widetilde{\mathbf{v}}_i=\mathbf{v}_i+g_i\mathbf{r}_i$, where $\mathbf{r}_i$ is adaptive latent noise. In this way, \textsc{Lens} achieves visual selectivity without cropping images, pruning tokens, generating intermediate images, or adding a persistent reasoning trace.
\textbf{Method Overview.}\quad As illustrated in Fig.~\ref{fig:motivation} and~\ref{fig:overview}, \textsc{Lens} aims to purify the visual context before decoding through two coupled stages, i.e., \textbf{evidence probing} and \textbf{latent suppression}. \textbf{Evidence probing} appends a temporary control token $\langle mask\rangle$ to the joint input $[\mathbf{V},\mathbf{T}]$. The hidden state of this token is decoded into a question-conditioned LET score vector $\mathbf{a}\in[0,1]^N$, where $a_i\in[0,1]$ measures the reliability of visual token $\mathbf{v}_i$ for the current question; the control token is removed afterwards. \textbf{Latent suppression} converts the LET scores into suppression gates $g_i$, which guide the construction of a purified visual sequence $\widetilde{\mathbf{V}}=\{\widetilde{\mathbf{v}}_i\}_{i=1}^{N}$ by forming $\widetilde{\mathbf{v}}_i=\mathbf{v}_i+g_i\mathbf{r}_i$, where $\mathbf{r}_i$ denotes adaptive latent noise. The final answer is then decoded as $P_{\psi}(Y \mid \widetilde{\mathbf{V}},\mathbf{T})$. In this way, \textsc{Lens} achieves visual selectivity while preserving the original visual token count and backbone interface, without cropping images, pruning tokens, generating intermediate images, or adding a persistent reasoning trace.

%\subsection{Question-Conditioned Evidence Probing}
\subsection{Evidence Probing}
\textsc{Lens} estimates token relevance through a short probing path. Given $\mathbf{V}$ and $\mathbf{T}$, we append the temporary \emph{Lens Evidence Token} $\langle mask\rangle$ after the text tokens and form
\begin{align}
\mathbf{S}_{\mathrm{probe}}
=
[\mathbf{v}_1,\ldots,\mathbf{v}_N,
\mathbf{t}_1,\ldots,\mathbf{t}_M,
\mathbf{p}_{m}], \label{eq:2}
\end{align}
where $\mathbf{p}_{m}$ represents the temporary $\langle mask\rangle$ control token in the joint MLLM input space. This token is appended only to the probing sequence and is not part of the original visual-token or text-token spaces.
%Here $\mathrm{MLLM}^{\mathrm{tok}}_{\psi}(\cdot)$ denotes the token-level MLLM backbone after the image has already been encoded into $\mathbf{V}$. It outputs a sequence of hidden states,
Then, deocoder $\mathrm{MLLM}^{\mathrm{tok}}_{\psi}(\cdot)$ outputs a sequence of hidden states based on $\mathbf{S}_{\mathrm{probe}}$,
\begin{align}
\mathbf{H}_{\mathrm{probe}}
&=
\mathrm{MLLM}^{\mathrm{tok}}_{\psi}(\mathbf{S}_{\mathrm{probe}}),\\
\mathbf{h}_{m}
&=
[\mathbf{H}_{\mathrm{probe}}]_{m}.
\end{align}
Since $\mathbf{h}_{m}$ attends to both visual and textual tokens, it provides a compact question-aware summary for evidence prediction. The LET only opens this probing path and is not used during final answer decoding.

A single-layer MLP head $f_{\theta}$ maps the LET hidden state to a dense evidence score vector over visual tokens
\begin{align}
\mathbf{a}
=
\sigma(f_{\theta}(\mathbf{h}_{m})),
\quad
\mathbf{a}\in[0,1]^N.
\end{align}
Each score $a_i$ indicates how likely $\mathbf{v}_i$ supports the answer. This LET score vector is conditioned on the question, so the same image region may be evidence for one question and a distractor for another.

We train the \emph{Lens Evidence Token} with object-level evidence supervision instead of chain-of-thought traces. Let $\mathcal{B}_{Q}$ denote the question-relevant boxes, and let $\Omega_i$ denote the image patch covered by visual token $\mathbf{v}_i$. The token label is
\begin{align}
z_i
=
\begin{cases}
1, & \exists B\in\mathcal{B}_{Q},\ \Omega_i\cap B\neq\emptyset,\\
0, & \mathrm{otherwise}.
\end{cases}
\end{align}
A visual token is labeled positive if its image patch overlaps at least one question-relevant box, and is labeled negative otherwise. The LET supervision loss is
\begin{align}
\mathcal{L}_{\mathrm{LET}}
=
-\frac{1}{N}
\sum_{i=1}^{N}
\left[
z_i\log a_i
+
(1-z_i)\log(1-a_i)
\right].
\end{align}
This supervision teaches the probe where visual evidence is located, while avoiding expensive rationale annotations.

%\subsection{LET-Guided Latent Suppression}
\subsection{Latent Suppression}
The LET scores become effective only when they change how visual tokens influence decoding. \textsc{Lens} therefore converts $\mathbf{a}$ into a latent reliability gate. The gate preserves tokens with high evidence scores and perturbs tokens with low evidence scores in the same latent space used by the MLLM.

\begin{table*}[tbp]
\renewcommand{\arraystretch}{1.0}
\centering
\setlength{\tabcolsep}{2.0pt}
\caption{Main comparison on 10 benchmarks for visual understanding, reasoning, and grounding. VQA scores evaluate answer quality, F1@0.5 evaluates grounding quality, and averages summarize VQA, grounding, and overall performance. The \textbf{best} and \underline{second best} values are marked.}
\label{tab:cmp}
% \vspace{-10pt}
\resizebox{1\linewidth}{!}{
\begin{tabular}{l|ccccccc|ccccc|c}
\toprule
\multirow{2}{*}{\textbf{Methods}} &
\multicolumn{7}{c|}{\textbf{General VQA Tasks}} &
\multicolumn{5}{c|}{\textbf{Grounding Tasks F1@0.5}} &
\multirow{2}{*}{\textbf{Avg.}}\\
\cmidrule(lr){2-8}\cmidrule(lr){9-13}
 & CUB & GQA & OpenImg & SROIE & VSR & MSVQA & Avg. & COCO & Obj365 & RUOD & Visdrone & Avg. \\
\midrule[0.5pt]
Vanilla & 68.60 & 64.06 & 50.00 & 88.71 & 66.54 & 50.41 & 64.72 & 12.88 & 9.12 & 9.42 & 20.44 & 12.97 & 44.02\\
\midrule[0.5pt]
SFT & 86.45 & 72.65 & 82.08 & 92.49 & 76.58 & 63.96 & 79.04 & 46.50 & 40.02 & 63.63 & 39.60 & 47.44 & 66.40\\
Visual-RFT & 88.21 & 68.95 & 73.74 & 94.28 & 72.28 & 62.14 & 76.60 & \underline{55.89} & 36.81 & 50.24 & \underline{46.32} & 47.32 & 64.89\\
VLM-R1 & 87.84 & 74.69 & 84.81 & 94.46 & 78.71 & 65.66 & 81.03 & 50.48 & 41.62 & 66.59 & 43.07 & 50.44 & 68.79\\
PAPO & 86.96 & 70.97 & 74.86 & 93.64 & 75.99 & 60.55 & 77.16 & 50.43 & 38.45 & 55.26 & 42.05 & 46.55 & 64.92\\
\midrule[0.5pt]
VPT &91.06 & 76.37 & 85.04 & 94.36 & 80.18 & 65.71 & 82.12 & 31.54 & 25.29 & 47.99 & 23.32 & 32.04 & 62.09\\
LVR & 90.28 & 72.61 & 78.68 & 90.45 & 76.93 & 63.50 & 78.74 & 49.77 & 39.89 & 64.78 & 42.34 & 49.20 & 66.92\\
DMLR & 90.26 & 75.56 & 85.13 & 88.41 & 78.45 & 64.28 & 80.35 & 44.39 & 38.67 & 58.79 & 35.28 & 44.28 & 65.92\\
Vismem & 89.77 & 71.44 & 80.58 & 90.67 & 75.45 & 62.78 & 78.45 & 45.29 & 39.25 & 62.16 & 37.67 & 46.09 & 65.51\\
\midrule[0.5pt]
\rowcolor{gray!15}
\textbf{\textsc{Lens-sft}} & \textbf{91.50} & \underline{79.01} & \underline{86.89} & \underline{95.20} & \underline{80.69} & \underline{66.39} & \underline{83.28} & 52.00 & \underline{44.10} & \underline{70.05} & 44.09 & \underline{52.56} & \underline{70.99}\\
\rowcolor{gray!15}
\textbf{\textsc{Lens-grpo}} & \underline{91.24} & \textbf{83.95} & \textbf{88.10} & \textbf{95.92} & \textbf{82.43} & \textbf{67.55} & \textbf{84.87} & \textbf{60.93} & \textbf{51.32} & \textbf{74.51} & \textbf{58.41} & \textbf{61.29} & \textbf{75.44}\\
\bottomrule
\end{tabular}
}
\end{table*}

For each visual token $\mathbf{v}_i\in\mathbb{R}^{d}$, a lightweight noise generator $G_{\phi}$ predicts the mean and positive scale of a token-specific noise distribution:
\begin{align}
(\boldsymbol{\mu}_i,\boldsymbol{\sigma}_i)
=
G_{\phi}(\mathbf{v}_i),
\quad
\boldsymbol{\sigma}_i>0.
\end{align}
We then sample $\boldsymbol{\epsilon}_i\sim\mathcal{N}(\mathbf{0},\mathbf{I})$ and obtain latent noise by reparameterization
\begin{align}
\mathbf{r}_i
=
\boldsymbol{\mu}_i
+
\boldsymbol{\sigma}_i\odot\boldsymbol{\epsilon}_i.
\end{align}
This makes the perturbation feature-aware rather than fixed random noise.

Given a threshold $\tau$, the LET-guided gate is $g_i = \frac{1}{\tau}\mathrm{ReLU}(\tau-a_i)$, where $0\leq g_i\leq 1$.
When $a_i\geq\tau$, the gate is zero and the token remains unchanged. When $a_i<\tau$, the gate increases as the evidence score decreases. The purified token is
\begin{align}
\widetilde{\mathbf{v}}_i
=
\mathbf{v}_i
+
g_i\mathbf{r}_i.
\end{align}
The resulting sequence $\widetilde{\mathbf{V}}=\{\widetilde{\mathbf{v}}_i\}_{i=1}^{N}$ keeps the original token layout. Thus the decoder still receives a full visual sequence, but low-reliability tokens become less stable as evidence for answer generation.

\subsection{Optimization and Inference}
%\textsc{Lens} is first trained with supervised fine-tuning. The answer loss is computed after latent suppression, so the model learns to answer from $\widetilde{\mathbf{V}}$. The prior loss grounds the evidence probe with token-level labels. The objective is
\textbf{Training.} \textsc{Lens} is first trained with supervised fine-tuning. The answer loss is computed after latent suppression, so the model learns to answer from $\widetilde{\mathbf{V}}$. The LET loss grounds the probe with token-level labels.
\begin{align}
\mathcal{L}_{\mathrm{SFT}}
=
\mathcal{L}_{\mathrm{ans}}
+
\beta\mathcal{L}_{\mathrm{LET}}.
\end{align}
Here, $\beta$ balances answer learning and LET supervision.
After supervised fine-tuning, we refine the evidence policy with reinforcement fine-tuning. The supervised LET loss trains token scores independently, while latent suppression depends on the selected evidence set. We model the LET score vector as a Bernoulli policy during training
\begin{align}
b_i \sim \mathrm{Bernoulli}(a_i),
\quad
\mathbf{b}=\{b_i\}_{i=1}^{N}.
\end{align}
The binary action $\mathbf{b}$ indicates selected evidence tokens. Given the label $\mathbf{z}$, the set-level reward is the F1 score
\begin{align}
R(\mathbf{b},\mathbf{z})
=
\frac{
2\cdot \sum_{i=1}^{N} b_i z_i
}{
\sum_{i=1}^{N} b_i
+
\sum_{i=1}^{N} z_i
+
\varepsilon
}.
\end{align}
The small constant $\varepsilon$ avoids numerical instability. We maximize the expected reward while regularizing the policy toward the supervised checkpoint
\begin{align}
\mathcal{J}_{\mathrm{RFT}}(\theta)
=
\mathbb{E}_{\mathbf{b}\sim\pi_{\theta}(\cdot\mid I,Q)}
\left[
R(\mathbf{b},\mathbf{z})
\right]
-
\lambda
D_{\mathrm{KL}}
\left(
\pi_{\theta}(\cdot\mid I,Q)
\|
\pi_{\mathrm{SFT}}(\cdot\mid I,Q)
\right).
\end{align}
%\textbf{Inference.} As illustrated in Fig.~\ref{fig:overview}, \textsc{Lens} first inserts the temporary $\langle mask\rangle$ and predicts LET scores $\mathbf{a}$. After removing the temporary token, it applies the threshold gate $g_i$ and forms $\widetilde{\mathbf{v}}_i=\mathbf{v}_i+g_i\mathbf{r}_i$ for each visual token, yielding the purified sequence $\widetilde{\mathbf{V}}$. The model then performs normal autoregressive decoding.
%
%During inference, \textsc{Lens} first inserts $\langle mask\rangle$ and predicts $\mathbf{a}$. It then removes the control token, constructs $\widetilde{\mathbf{V}}$ with LFT-guided latent suppression, and performs normal autoregressive decoding. The generation sequence is
%\begin{align}
%\mathbf{S}_{\mathrm{gen}}
%=
%[\widetilde{\mathbf{v}}_1,\ldots,\widetilde{\mathbf{v}}_N,
%\mathbf{t}_1,\ldots,\mathbf{t}_M],
%\end{align}
%and the answer is sampled from
%\begin{align}
%Y
%\sim
%P_{\psi}(\cdot\mid\mathbf{S}_{\mathrm{gen}}).
%\end{align}
%This inference process uses no external detector, no image generation, and no test-time latent optimization. The only extra cost is the probing pass and the latent suppression operation.
\textbf{Inference.}\quad During inference, \textsc{Lens} first inserts the temporary $\langle mask\rangle$ and predicts LET scores $\mathbf{a}$. It then removes the control token, applies the threshold gate $g_i$ and forms $\widetilde{\mathbf{v}}_i=\mathbf{v}_i+g_i\mathbf{r}_i$ for each visual token, yielding the purified sequence $\widetilde{\mathbf{V}}$ with LFT-guided latent suppression. The generation sequence is
\begin{align}
\mathbf{S}_{\mathrm{gen}}
=
[\widetilde{\mathbf{v}}_1,\ldots,\widetilde{\mathbf{v}}_N,
\mathbf{t}_1,\ldots,\mathbf{t}_M],
\end{align}
and the answer is sampled from
\begin{align}
Y
\sim
P_{\psi}(\cdot\mid\mathbf{S}_{\mathrm{gen}}).
\end{align}
This inference process uses no external detector, no image generation, and no test-time latent optimization. The only extra cost is the probing pass and the latent suppression operation.

\begin{table*}[tbp]
\renewcommand{\arraystretch}{0.9}
\centering
\setlength{\tabcolsep}{1.6pt}
\newcommand{\gain}[1]{\textcolor{green!50!black}{\scriptsize$\uparrow$#1}}
\newcommand{\drop}[1]{\textcolor{red!65!black}{\scriptsize$\downarrow$#1}}
\caption{Generalization results across 9 base models from Qwen3-VL~\citep{qwen3vl}, InternVL3.5-VL~\citep{internvl3_5}, and Qwen3.5~\citep{qwen3_5}. Green $\uparrow$ and red $\downarrow$ values denote absolute changes from the corresponding base model, showing how the same visual evidence purification strategy transfers across model families and scales.}
\label{tab:basemodels}
\resizebox{1\linewidth}{!}{
\begin{tabular}{l|*{6}{r@{\,}l}|*{4}{r@{\,}l}|r@{\,}l}
\toprule
\textbf{Base Model} &
\multicolumn{2}{|l}{\textbf{CUB}} &
\multicolumn{2}{l}{\textbf{GQA}} &
\multicolumn{2}{l}{\textbf{OpenImg}} &
\multicolumn{2}{l}{\textbf{SROIE}} &
\multicolumn{2}{l}{\textbf{VSR}} &
\multicolumn{2}{l|}{\textbf{MSVQA}} &
\multicolumn{2}{l}{\textbf{COCO}} &
\multicolumn{2}{l}{\textbf{Obj365}} &
\multicolumn{2}{l}{\textbf{RUOD}} &
\multicolumn{2}{l|}{\textbf{Visdrone}} &
\multicolumn{2}{l}{\textbf{Avg.}}\\
\midrule[0.5pt]
Qwen3-VL-2B & 83.5 &  & 70.5 &  & 82.8 &  & 93.8 &  & 74.3 &  & 62.8 &  & 44.0 &  & 36.9 &  & 62.5 &  & 32.4 &  & 64.3 & \\
\rowcolor{gray!15}
\textbf{+ \textsc{Lens-sft}} & 88.1 & \gain{4.6} & 78.4 & \gain{7.9} & 86.7 & \gain{3.8} & 94.7 & \gain{0.9} & 78.7 & \gain{4.5} & 65.0 & \gain{2.2} & 47.8 & \gain{3.8} & 39.5 & \gain{2.7} & 65.9 & \gain{3.4} & 35.0 & \gain{2.5} & 68.0 & \gain{3.6}\\
\rowcolor{gray!15}
\textbf{+ \textsc{Lens-grpo}} & 89.6 & \gain{6.2} & 78.7 & \gain{8.2} & 87.5 & \gain{4.6} & 94.3 & \gain{0.5} & 79.7 & \gain{5.5} & 68.0 & \gain{5.2} & 52.5 & \gain{8.5} & 42.0 & \gain{5.2} & 66.9 & \gain{4.4} & 44.3 & \gain{11.8} & 70.3 & \gain{6.0}\\
Qwen3-VL-4B & 86.5 &  & 72.7 &  & 82.1 &  & 92.5 &  & 76.6 &  & 64.0 &  & 46.5 &  & 40.0 &  & 63.6 &  & 39.6 &  & 66.4 & \\
\rowcolor{gray!15}
\textbf{+ \textsc{Lens-sft}} & 91.5 & \gain{5.1} & 79.0 & \gain{6.4} & 86.9 & \gain{4.8} & 95.2 & \gain{2.7} & 80.7 & \gain{4.1} & 66.4 & \gain{2.4} & 52.0 & \gain{5.5} & 44.1 & \gain{4.1} & 70.1 & \gain{6.4} & 44.1 & \gain{4.5} & 71.0 & \gain{4.6}\\
\rowcolor{gray!15}
\textbf{+ \textsc{Lens-grpo}} & 91.2 & \gain{4.8} & 84.0 & \gain{11.3} & 88.1 & \gain{6.0} & \textbf{95.9} & \gain{3.4} & \textbf{82.4} & \gain{5.9} & 67.6 & \gain{3.6} & \textbf{60.9} & \gain{14.4} & 51.3 & \gain{11.3} & 74.5 & \gain{10.9} & 58.4 & \gain{18.8} & 75.4 & \gain{9.0}\\
Qwen3-VL-8B & 89.5 &  & 72.2 &  & 85.2 &  & 94.7 &  & 76.2 &  & 65.1 &  & 49.9 &  & 41.2 &  & 68.2 &  & 42.3 &  & 68.4 & \\
\rowcolor{gray!15}
\textbf{+ \textsc{Lens-sft}} & \textbf{92.3} & \gain{2.8} & 79.6 & \gain{7.4} & 87.4 & \gain{2.2} & 95.7 & \gain{1.0} & 79.2 & \gain{3.0} & 70.2 & \gain{5.1} & 52.2 & \gain{2.3} & 43.3 & \gain{2.1} & 72.9 & \gain{4.7} & 45.7 & \gain{3.5} & 71.8 & \gain{3.4}\\
\rowcolor{gray!15}
\textbf{+ \textsc{Lens-grpo}} & 91.7 & \gain{2.2} & \textbf{82.4} & \gain{10.2} & \textbf{88.3} & \gain{3.0} & 95.5 & \gain{0.8} & 81.8 & \gain{5.6} & \textbf{70.9} & \gain{5.9} & 60.0 & \gain{10.1} & \textbf{52.7} & \gain{11.4} & \textbf{75.6} & \gain{7.4} & \textbf{60.2} & \gain{17.9} & \textbf{75.9} & \gain{7.4}\\
\midrule[0.5pt]
InternVL3.5-2B & 86.9 &  & 73.9 &  & 84.1 &  & 92.1 &  & 76.5 &  & 66.0 &  & 33.0 &  & 22.7 &  & 38.5 &  & 17.2 &  & 59.1 & \\
\rowcolor{gray!15}
\textbf{+ \textsc{Lens-sft}} & 90.5 & \gain{3.6} & 83.7 & \gain{9.8} & 86.8 & \gain{2.7} & 94.1 & \gain{2.0} & 78.3 & \gain{1.8} & 68.2 & \gain{2.2} & 49.7 & \gain{16.7} & 36.3 & \gain{13.6} & 56.5 & \gain{18.0} & 29.3 & \gain{12.1} & 67.3 & \gain{8.2}\\
InternVL3.5-4B & 89.7 &  & 74.7 &  & 84.8 &  & 93.3 &  & 77.0 &  & 66.7 &  & 47.6 &  & 37.9 &  & 53.6 &  & 28.5 &  & 65.4 & \\
\rowcolor{gray!15}
\textbf{+ \textsc{Lens-sft}} & 90.9 & \gain{1.2} & 87.5 & \gain{12.8} & 86.3 & \gain{1.5} & 95.3 & \gain{2.0} & \textbf{83.0} & \gain{6.0} & 70.5 & \gain{3.8} & 53.7 & \gain{6.1} & \textbf{45.1} & \gain{7.2} & 64.7 & \gain{11.1} & 37.2 & \gain{8.7} & 71.4 & \gain{6.0}\\
InternVL3.5-8B & 91.3 &  & 77.6 &  & 86.2 &  & 93.8 &  & 80.7 &  & 68.9 &  & 48.4 &  & 39.5 &  & 55.3 &  & 29.9 &  & 67.1 & \\
\rowcolor{gray!15}
\textbf{+ \textsc{Lens-sft}} & \textbf{93.1} & \gain{1.8} & \textbf{90.1} & \gain{12.5} & \textbf{87.3} & \gain{1.1} & \textbf{96.0} & \gain{2.2} & 82.5 & \gain{1.8} & \textbf{71.5} & \gain{2.6} & \textbf{54.0} & \gain{5.6} & 44.6 & \gain{5.1} & \textbf{66.9} & \gain{11.6} & \textbf{37.4} & \gain{7.5} & \textbf{72.3} & \gain{6.9}\\
\midrule[0.5pt]
Qwen3.5-2B & 86.2 &  & 72.7 &  & 84.5 &  & 92.7 &  & 76.7 &  & 65.9 &  & 38.4 &  & 34.9 &  & 59.0 &  & 36.4 &  & 64.7 & \\
\rowcolor{gray!15}
\textbf{+ \textsc{Lens-sft}} & 89.1 & \gain{2.9} & 77.6 & \gain{4.9} & 88.3 & \gain{3.8} & 93.7 & \gain{1.0} & 78.1 & \gain{1.4} & 69.0 & \gain{3.1} & 43.2 & \gain{4.8} & 38.8 & \gain{3.9} & 63.6 & \gain{4.6} & 37.3 & \gain{0.9} & 67.8 & \gain{3.1}\\
Qwen3.5-4B & 89.8 &  & 79.9 &  & 87.0 &  & 94.7 &  & 80.2 &  & 67.5 &  & 41.3 &  & 38.9 &  & 64.5 &  & 39.1 &  & 68.3 & \\
\rowcolor{gray!15}
\textbf{+ \textsc{Lens-sft}} & 91.3 & \gain{1.6} & 89.2 & \gain{9.3} & \textbf{90.4} & \gain{3.4} & 94.5 & \drop{0.2} & 81.4 & \gain{1.2} & 69.4 & \gain{1.8} & 48.2 & \gain{6.8} & 42.3 & \gain{3.4} & 65.9 & \gain{1.5} & 40.6 & \gain{1.6} & 71.3 & \gain{3.0}\\
Qwen3.5-9B & 91.3 &  & 79.6 &  & 84.5 &  & \textbf{95.0} &  & 81.7 &  & 69.7 &  & 44.5 &  & 42.2 &  & 66.4 &  & 42.3 &  & 69.7 & \\
\rowcolor{gray!15}
\textbf{+ \textsc{Lens-sft}} & \textbf{93.3} & \gain{2.0} & \textbf{90.9} & \gain{11.3} & 88.3 & \gain{3.8} & 94.7 & \drop{0.3} & \textbf{82.4} & \gain{0.7} & \textbf{71.0} & \gain{1.3} & \textbf{52.7} & \gain{8.2} & \textbf{45.4} & \gain{3.2} & \textbf{68.4} & \gain{2.0} & \textbf{44.5} & \gain{2.2} & \textbf{73.1} & \gain{3.4}\\
\bottomrule
\end{tabular}
}
\end{table*}

\section{Experiments}
\subsection{Experimental Settings}
\paragraph{Benchmarks.}
We evaluate \textsc{Lens} on 10 datasets covering VQA and grounding. For VQA, we use CUB~\citep{cub}, GQA~\citep{gqa}, OpenImages~\citep{openimages}, SROIE~\citep{sroie}, VSR~\citep{vsr}, and MSVQA~\citep{msvqa}. For grounding, we use COCO2017~\citep{coco2017}, Objects365~\citep{objects365}, RUOD~\citep{ruod}, and VisDrone~\citep{visdrone}. We report VQA scores following the Visual CoT benchmark~\citep{shao2024visual}. For grounding, we report F1@0.5 following object detection metrics. More dataset details are provided in Appendix~\ref{appdatasets}.

\paragraph{Baselines.}
We compare \textsc{Lens} with Vanilla and SFT, as well as 7 representative reasoning and perception baselines. These include Visual-RFT~\citep{liu2025visual}, VLM-R1~\citep{shen2025vlm}, PAPO~\citep{wang2025perception}, VPT~\citep{yu2025introducing}, LVR, DMLR~\citep{liu2025reasoning}, and VisMem~\citep{yu2025vismem}.

\paragraph{Implementation Details.}
All experiments except Table~\ref{tab:basemodels} use Qwen3-VL-4B and 8 NVIDIA H100 80G GPUs. We set $\beta$ to 0.2 to balance the two loss terms and set $\tau$ to 0.5 corresponding to the binary cross-entropy loss. More implementation details are provided in Appendix~\ref{appimplementation}.

\begin{figure*}[!t]
    \centering
    \includegraphics[width=5.3in]{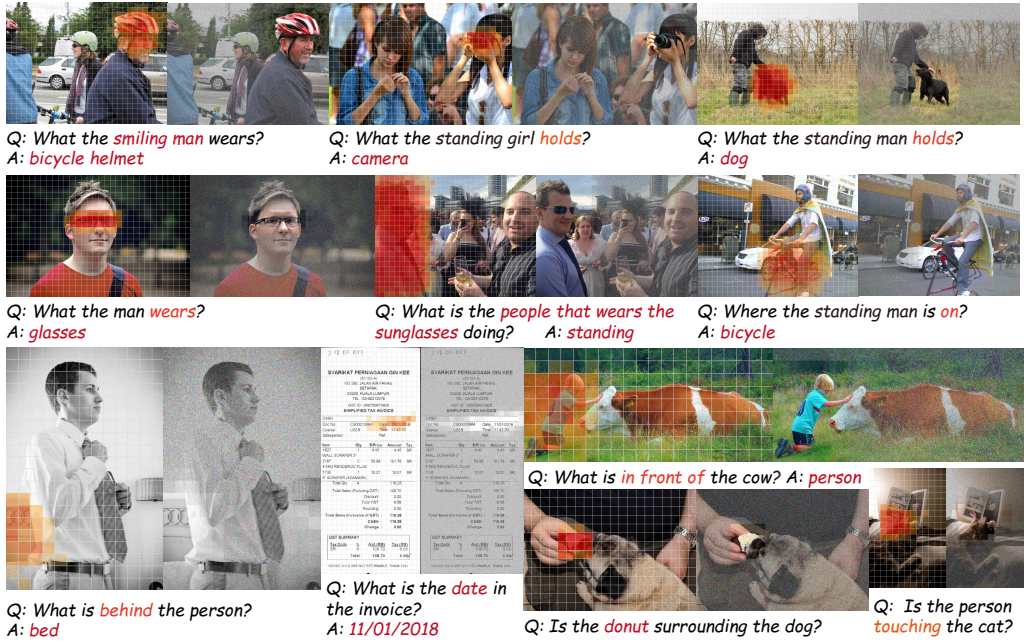}
%    \caption{Visualization of \textsc{Lens} on VQA tasks. The predicted LET score vector is mapped back to image patches, where warmer overlays indicate higher question relevance. The examples show that \textsc{Lens} focuses on answer-supporting regions and reduces distractor activation.} \label{fig:vis_1}
	\caption{Visualization of \textsc{Lens} on VQA tasks. For each example, the left image shows the question-conditioned LET scores mapped to visual patches, and the right image shows the visual effect after applying token-level latent noise to low-relevance visual tokens. \textsc{Lens} preserves answer-supporting evidence such as attributes, objects, relations, and OCR fields, while suppressing irrelevant background or distracting regions. This illustrates its advantage in purifying visual evidence for fine-grained answer generation.} \label{fig:vis_1}
\end{figure*}

\subsection{Main Results}
\textbf{The main comparison in Table~\ref{tab:cmp} shows that \textsc{Lens} improves MLLM performance by cleaning visual evidence rather than adding more reasoning context.} Compared with SFT, \textsc{Lens-sft} raises the overall average from 66.40 to 70.99, and \textsc{Lens-grpo} further increases it to 75.44. Against the strongest non-\textsc{Lens} baseline, \textsc{Lens-grpo} improves the overall average by 6.65 points. This gain is not concentrated in a single benchmark group. It improves the VQA average from 79.04 to 84.87 and the grounding average from 47.44 to 61.29 over SFT, showing that evidence purification benefits both answer prediction and spatial localization.

\textbf{The VQA results indicate that suppressing redundant visual tokens improves fine-grained answer generation while preserving broad multimodal reasoning ability.} \textsc{Lens-sft} improves all six VQA datasets over SFT, with gains from 2.43 points on MSVQA to 6.36 points on GQA. The reinforcement refinement further raises the VQA average to 84.87 and obtains the best results on five of the six VQA datasets. The largest improvement appears on GQA, where \textsc{Lens-grpo} reaches 83.95 compared with 72.65 for SFT. This pattern suggests that the learned LET scores help the model focus on evidence needed for compositional and fine-grained reasoning.

\textbf{The grounding results provide stronger evidence that visual selectivity is the main source of improvement.} Grounding requires the model to locate question-relevant regions, so distractor suppression should directly improve this metric. \textsc{Lens-sft} increases the grounding average from 47.44 to 52.56 over SFT, while \textsc{Lens-grpo} raises it to 61.29. The gains of \textsc{Lens-grpo} are especially large on VisDrone, COCO, Object365, and RUOD, reaching 18.81, 14.43, 11.30, and 10.88 points over SFT. These improvements support the claim that latent noise masking reduces the influence of irrelevant visual tokens rather than merely increasing model capacity.

\textbf{Table~\ref{tab:basemodels} shows that \textsc{Lens} is a transferable visual evidence purification strategy across model families and scales.} With supervised training alone, \textsc{Lens-sft} improves the overall average on all nine base models by 3.0--8.2 points. The gains hold for Qwen3-VL, InternVL3.5-VL, and Qwen3.5 models, indicating that the intervention is not tied to one backbone. On Qwen3-VL models, reinforcement refinement further enlarges the overall gains to 6.0, 9.0, and 7.4 points for 2B, 4B, and 8B models. Although Qwen3.5 shows two small drops on SROIE, the overall averages remain consistently higher, suggesting that \textsc{Lens} improves visual selectivity with limited task-specific tradeoffs.

\begin{figure*}[!t]
    \centering
    \includegraphics[width=5.3in]{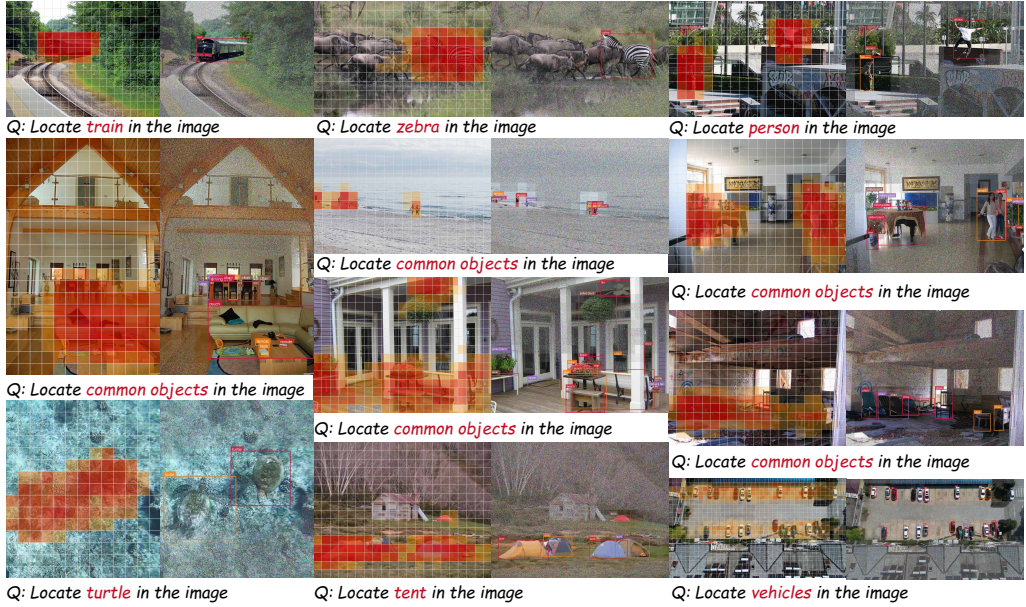}
%    \caption{Visualization of the temporary $\langle mask\rangle$ control token in \textsc{Lens} on grounding tasks. Its predicted token-level evidence gate is mapped back to image patches, where warmer overlays indicate higher question relevance. The examples show that this control token assigns higher gate scores to target objects and lower scores to surrounding distractors.} \label{fig:vis_2}
	\caption{Visualization of \textsc{Lens} on grounding tasks. For each example, the left image shows the LET scores over visual patches, and the right image shows the visual effect after applying token-level latent noise to low-relevance visual tokens. \textsc{Lens} keeps target objects and multiple queried instances visually stable, while perturbing surrounding clutter and non-target distractors. This illustrates its advantage in improving question-conditioned spatial localization.} \label{fig:vis_2}
\vspace{-10pt}
\end{figure*}

\subsection{Ablation Study and Visualizations}
\subsubsection{Ablation Study}
\textbf{The ablation results in Table~\ref{tab:ablation} show that \textsc{Lens} works because evidence prediction and latent suppression act together.} Evidence probing alone improves the average score from 66.40 to 68.04, which indicates that the temporary $\langle mask\rangle$ token learns a useful question-conditioned LET scores. However, this gain is limited because the LFT only identifies relevant regions and does not yet reduce the influence of redundant tokens during decoding. This supports the main claim that visual selectivity must be connected to an explicit latent intervention.

\textbf{Different masking types confirm that visual suppression must be guided by the learned LET scores and soft noise.} EP+RM drops the average score to 60.95, and the degradation is especially clear on tasks that require precise evidence, e.g. SROIE and GQA. It shows that simply perturbing visual tokens can destroy answer-supporting cues. In addition, EP+ZM/AM replaces the low-score token with zero padding or average visual token, which also drop the average score to 58.66/59.67, respectively. In contrast, EP+LS reaches 70.99 and improves all benchmarks over the baseline, showing that latent suppression with soft noise is effective when it preserves high-score tokens and weakens low-score distractors.

\textbf{LET-guided visual suppression significantly outperforms applying noise based on attention scores.} \textit{Attention mask} drops the average score to 65.33. We suspect that the attention of a certain layer does not always focus on direct visual evidence, so applying visual suppression based on the attention scores of that layer is inadvisable. Moreover, this approach requires obtaining the complete attention weights, making it impossible to use acceleration algorithms (such as flash attention), which significantly reduces training and inference efficiency. This indicates that the proposed method has a clear advantage over applying visual suppression based on model attention.

\textbf{The reinforcement results further show that set-level evidence quality is crucial for the final visual context.} SFT+GRPO raises the average score from 66.40 to 70.64, showing that policy optimization improves the base model. When the same optimization is applied after \textsc{Lens-sft}, the average rises to 73.57, which shows that a cleaner LET scoring signal provides a stronger starting point. \textsc{Lens-grpo} obtains the best average score of 75.44, indicating that optimizing the evidence policy with a set-level reward better matches the goal of selecting complete question-relevant evidence. \textbf{These ablations also clarify why \textsc{Lens} improves both VQA and grounding.} The LET scores alone brings moderate gains because it exposes question-relevant regions, while latent suppression turns these scores into a cleaner decoding context. The large drop of EP+RM shows that suppression is useful only when it follows evidence structure. The stronger result of \textsc{Lens-grpo} further suggests that complete evidence sets matter more than isolated high-score patches.

\begin{table*}[!t]
\renewcommand{\arraystretch}{0.9}
\centering
\caption{Ablation on Qwen3-VL-4B evaluating the role of evidence probing, random masking, latent suppression, and reinforcement refinement in visual evidence purification. EP denotes \textit{Evidence probing}, RM denotes \textit{Random mask}, Attention mask directly uses the attention scores of the last layer of the decorder to add noise to the visual token sequence, ZM replaces low-score tokens with zero padding, AM replaces them with average visual tokens, and LS denotes \textit{Latent suppression}.}
\label{tab:ablation}
\resizebox{1\linewidth}{!}{
\begin{tabular}{l|cccccc|cccc|c}
\toprule
Methods & CUB & GQA & OpenImg & SROIE & VSR & MSVQA & COCO & Obj365 & RUOD & Visdrone & Avg.\\ 
\midrule[0.5pt]
Baselines & 86.45 & 72.65 & 82.08 & 92.49 & 76.58 & 63.96 & 46.50 & 40.02 & 63.63 & 39.60 & 66.40\\
\midrule[0.5pt]
EP & 87.62 & 74.89 & 83.65 & 93.15 & 77.67 & 64.36 & 48.98 & 41.85 & 66.52 & 41.67 & 68.04\\
EP+RM & 85.60 & 68.34 & 84.13 & 69.90 & 70.67 & 60.48 & 41.47 & 34.52 & 60.73 & 33.70 & 60.95\\
EP+ZM & 87.23 & 63.99 & 82.38 & 56.43 & 69.55 & 61.41 & 43.02 & 36.84 & 56.28 & 29.44 & 58.66\\
EP+AM & 86.82 & 65.47 & 83.23 & 59.88 & 68.19 & 62.68 & 44.91 & 36.62 & 58.97 & 29.92 & 59.67\\
Attention mask & 87.63 & 69.31 & 83.97 & 92.42 & 72.28 & 64.90 & 46.48 & 39.07 & 58.96 & 38.31 & 65.33 \\
\rowcolor{gray!15}
EP+LS & \textbf{91.50} & \textbf{79.01} & \textbf{86.89} & \textbf{95.20} & \textbf{80.69} & \textbf{66.39} & \textbf{52.00} & \textbf{44.10} & \textbf{70.05} & \textbf{44.09} & \textbf{70.99}\\
\midrule[0.5pt]
SFT+GRPO & 90.85 & 76.69 & 85.03 & 94.62 & 78.71 & 67.25 & 52.75 & 46.96 & 65.21 & 48.36 & 70.64\\
EP+LS+GRPO & 91.72 & 81.65 & 87.20 & 95.13 & 79.35 & 67.32 & 56.34 & 48.46 & 71.96 & 56.58 & 73.57\\
\rowcolor{gray!15}
\textsc{Lens-grpo} & \textbf{91.24} & \textbf{83.95} & \textbf{88.10} & \textbf{95.92} & \textbf{82.43} & \textbf{67.55} & \textbf{60.93} & \textbf{51.32} & \textbf{74.51} & \textbf{58.41} & \textbf{75.44}\\
\bottomrule
\end{tabular}
}
\vspace{-10pt}
\end{table*}

\subsubsection{Visualizations}
Fig.~\ref{fig:vis_1} and Fig.~\ref{fig:vis_2} visualize the effect of LENS on VQA and grounding tasks. In each pair, the left image maps the predicted LET scores to visual patches, while the right image shows the image-space effect of applying token-level latent noise to low-relevance visual tokens. Since the intervention is performed in latent space, the right image should be interpreted as an effect visualization rather than pixel-level noise injection. For VQA tasks (Fig.~\ref{fig:vis_1}), LENS highlights answer-supporting cues such as attributes, objects, relations, and OCR fields, while the latent-noise effect mainly appears on backgrounds or unrelated regions. This shows its advantage in purifying fine-grained visual evidence for answer generation. For grounding tasks (Fig.~\ref{fig:vis_2}), LENS preserves target objects and multiple queried instances, while perturbing surrounding clutter and non-target distractors. This shows its advantage in improving question-conditioned spatial localization. Together, the visualizations support the claim that LENS improves multimodal reasoning by cleaning existing visual tokens rather than adding extra visual or textual context.

\section{Conclusion}

\textbf{This paper presents \textsc{Lens}, a visual evidence purification framework that improves multimodal reasoning by suppressing question-irrelevant visual tokens in latent space.} Instead of adding longer textual traces, extra visual inputs, or persistent latent memories, \textsc{Lens} introduces a question-conditioned \emph{Lens Evidence Token} supervised by object-level annotations and uses adaptive latent noise to weaken low-relevance tokens while preserving the original backbone and token sequence. Experiments across VQA and grounding benchmarks show consistent gains over strong training, token-level, and latent reasoning baselines. Ablations and visualizations further confirm that the improvement comes from coupling evidence probing with LET-guided latent suppression. These results suggest that cleaner visual evidence is a direct and effective path toward more reliable fine-grained vision-language reasoning.

\bibliography{iclr2026_conference}

@String(AAAI = {AAAI})

@ARTICLE{rne,
  author={Jiang, Kai and Bai, Xueru and Zhou, Feng},
  journal={IEEE Transactions on Neural Networks and Learning Systems}, 
  title={Recurrent Network Expansion for Class Incremental Learning}, 
  year={2026},
  volume={37},
  number={1},
  pages={122-135},
  doi={10.1109/TNNLS.2025.3601373}
}

@article{qwen3vl,
  title={Qwen3-vl technical report},
  author={Bai, Shuai and Cai, Yuxuan and Chen, Ruizhe and Chen, Keqin and Chen, Xionghui and Cheng, Zesen and Deng, Lianghao and Ding, Wei and Gao, Chang and Ge, Chunjiang and others},
  journal={arXiv preprint arXiv:2511.21631},
  year={2025}
}

@article{llava-ov,
  title={Llava-onevision: Easy visual task transfer},
  author={Li, Bo and Zhang, Yuanhan and Guo, Dong and Zhang, Renrui and Li, Feng and Zhang, Hao and Zhang, Kaichen and Zhang, Peiyuan and Li, Yanwei and Liu, Ziwei and others},
  journal={arXiv preprint arXiv:2408.03326},
  year={2024}
}

@article{internvl3_5,
  title={Internvl3. 5: Advancing open-source multimodal models in versatility, reasoning, and efficiency},
  author={Wang, Weiyun and Gao, Zhangwei and Gu, Lixin and Pu, Hengjun and Cui, Long and Wei, Xingguang and Liu, Zhaoyang and Jing, Linglin and Ye, Shenglong and Shao, Jie and others},
  journal={arXiv preprint arXiv:2508.18265},
  year={2025}
}

@inproceedings{zhang2024mathverse,
  title={Mathverse: Does your multi-modal llm truly see the diagrams in visual math problems?},
  author={Zhang, Renrui and Jiang, Dongzhi and Zhang, Yichi and Lin, Haokun and Guo, Ziyu and Qiu, Pengshuo and Zhou, Aojun and Lu, Pan and Chang, Kai-Wei and Qiao, Yu and others},
  booktitle={European Conference on Computer Vision},
  pages={169--186},
  year={2024},
  organization={Springer}
}

@article{wang2025perception,
  title={Perception-aware policy optimization for multimodal reasoning},
  author={Wang, Zhenhailong and Guo, Xuehang and Stoica, Sofia and Xu, Haiyang and Wang, Hongru and Ha, Hyeonjeong and Chen, Xiusi and Chen, Yangyi and Yan, Ming and Huang, Fei and others},
  journal={arXiv preprint arXiv:2507.06448},
  year={2025}
}

@article{wei2022chain,
  title={Chain-of-thought prompting elicits reasoning in large language models},
  author={Wei, Jason and Wang, Xuezhi and Schuurmans, Dale and Bosma, Maarten and Xia, Fei and Chi, Ed and Le, Quoc V and Zhou, Denny and others},
  journal={Advances in neural information processing systems},
  volume={35},
  pages={24824--24837},
  year={2022}
}

@article{zhang2023multimodal,
  title={Multimodal chain-of-thought reasoning in language models},
  author={Zhang, Zhuosheng and Zhang, Aston and Li, Mu and Zhao, Hai and Karypis, George and Smola, Alex},
  journal={arXiv preprint arXiv:2302.00923},
  year={2023}
}

@article{shao2024visual,
  title={Visual cot: Advancing multi-modal language models with a comprehensive dataset and benchmark for chain-of-thought reasoning},
  author={Shao, Hao and Qian, Shengju and Xiao, Han and Song, Guanglu and Zong, Zhuofan and Wang, Letian and Liu, Yu and Li, Hongsheng},
  journal={Advances in Neural Information Processing Systems},
  volume={37},
  pages={8612--8642},
  year={2024}
}

@article{zheng2025deepeyes,
  title={Deepeyes: Incentivizing" thinking with images" via reinforcement learning},
  author={Zheng, Ziwei and Yang, Michael and Hong, Jack and Zhao, Chenxiao and Xu, Guohai and Yang, Le and Shen, Chao and Yu, Xing},
  journal={arXiv preprint arXiv:2505.14362},
  year={2025}
}

@article{su2025openthinkimg,
  title={Openthinkimg: Learning to think with images via visual tool reinforcement learning},
  author={Su, Zhaochen and Li, Linjie and Song, Mingyang and Hao, Yunzhuo and Yang, Zhengyuan and Zhang, Jun and Chen, Guanjie and Gu, Jiawei and Li, Juntao and Qu, Xiaoye and others},
  journal={arXiv preprint arXiv:2505.08617},
  year={2025}
}

@article{yang2025machine,
  title={Machine mental imagery: Empower multimodal reasoning with latent visual tokens},
  author={Yang, Zeyuan and Yu, Xueyang and Chen, Delin and Shen, Maohao and Gan, Chuang},
  journal={arXiv preprint arXiv:2506.17218},
  year={2025}
}

@article{liu2025reasoning,
  title={Reasoning within the mind: Dynamic multimodal interleaving in latent space},
  author={Liu, Chengzhi and Yang, Yuzhe and Fan, Yue and Wei, Qingyue and Liu, Sheng and Wang, Xin Eric},
  journal={arXiv preprint arXiv:2512.12623},
  year={2025}
}

@article{yu2025vismem,
  title={Vismem: Latent vision memory unlocks potential of vision-language models},
  author={Yu, Xinlei and Xu, Chengming and Zhang, Guibin and Chen, Zhangquan and Zhang, Yudong and He, Yongbo and Jiang, Peng-Tao and Zhang, Jiangning and Hu, Xiaobin and Yan, Shuicheng},
  journal={arXiv preprint arXiv:2511.11007},
  year={2025}
}

@inproceedings{mondal2024kam,
  title={Kam-cot: Knowledge augmented multimodal chain-of-thoughts reasoning},
  author={Mondal, Debjyoti and Modi, Suraj and Panda, Subhadarshi and Singh, Rituraj and Rao, Godawari Sudhakar},
  booktitle={Proceedings of the AAAI conference on artificial intelligence},
  volume={38},
  number={17},
  pages={18798--18806},
  year={2024}
}

@inproceedings{gao2025interleaved,
  title={Interleaved-modal chain-of-thought},
  author={Gao, Jun and Li, Yongqi and Cao, Ziqiang and Li, Wenjie},
  booktitle={Proceedings of the Computer Vision and Pattern Recognition Conference},
  pages={19520--19529},
  year={2025}
}

@inproceedings{mitra2024compositional,
  title={Compositional chain-of-thought prompting for large multimodal models},
  author={Mitra, Chancharik and Huang, Brandon and Darrell, Trevor and Herzig, Roei},
  booktitle={Proceedings of the IEEE/CVF Conference on Computer Vision and Pattern Recognition},
  pages={14420--14431},
  year={2024}
}

@article{hu2024visual,
  title={Visual sketchpad: Sketching as a visual chain of thought for multimodal language models},
  author={Hu, Yushi and Shi, Weijia and Fu, Xingyu and Roth, Dan and Ostendorf, Mari and Zettlemoyer, Luke and Smith, Noah A and Krishna, Ranjay},
  journal={Advances in Neural Information Processing Systems},
  volume={37},
  pages={139348--139379},
  year={2024}
}

@article{fu2025refocus,
  title={Refocus: Visual editing as a chain of thought for structured image understanding},
  author={Fu, Xingyu and Liu, Minqian and Yang, Zhengyuan and Corring, John and Lu, Yijuan and Yang, Jianwei and Roth, Dan and Florencio, Dinei and Zhang, Cha},
  journal={arXiv preprint arXiv:2501.05452},
  year={2025}
}

@article{li2025imagine,
  title={Imagine while reasoning in space: Multimodal visualization-of-thought},
  author={Li, Chengzu and Wu, Wenshan and Zhang, Huanyu and Xia, Yan and Mao, Shaoguang and Dong, Li and Vuli{\'c}, Ivan and Wei, Furu},
  journal={arXiv preprint arXiv:2501.07542},
  year={2025}
}

@article{zhao2025pyvision,
  title={Pyvision: Agentic vision with dynamic tooling},
  author={Zhao, Shitian and Zhang, Haoquan and Lin, Shaoheng and Li, Ming and Wu, Qilong and Zhang, Kaipeng and Wei, Chen},
  journal={arXiv preprint arXiv:2507.07998},
  year={2025}
}

@article{wang2025pixel,
  title={Pixel reasoner: Incentivizing pixel-space reasoning with curiosity-driven reinforcement learning},
  author={Wang, Haozhe and Su, Alex and Ren, Weiming and Lin, Fangzhen and Chen, Wenhu},
  journal={arXiv preprint arXiv:2505.15966},
  year={2025}
}

@article{fan2025grit,
  title={Grit: Teaching mllms to think with images},
  author={Fan, Yue and He, Xuehai and Yang, Diji and Zheng, Kaizhi and Kuo, Ching-Chen and Zheng, Yuting and Narayanaraju, Sravana Jyothi and Guan, Xinze and Wang, Xin Eric},
  journal={arXiv preprint arXiv:2505.15879},
  year={2025}
}

@article{hao2024training,
  title={Training large language models to reason in a continuous latent space},
  author={Hao, Shibo and Sukhbaatar, Sainbayar and Su, DiJia and Li, Xian and Hu, Zhiting and Weston, Jason and Tian, Yuandong},
  journal={arXiv preprint arXiv:2412.06769},
  year={2024}
}

@article{li2025seek,
  title={Seek in the dark: Reasoning via test-time instance-level policy gradient in latent space},
  author={Li, Hengli and Li, Chenxi and Wu, Tong and Zhu, Xuekai and Wang, Yuxuan and Yu, Zhaoxin and Jiang, Eric Hanchen and Zhu, Song-Chun and Jia, Zixia and Wu, Ying Nian and others},
  journal={arXiv preprint arXiv:2505.13308},
  year={2025}
}

@article{qin2025chain,
  title={Chain-of-visual-thought: Teaching vlms to see and think better with continuous visual tokens},
  author={Qin, Yiming and Wei, Bomin and Ge, Jiaxin and Kallidromitis, Konstantinos and Fu, Stephanie and Darrell, Trevor and Wang, XuDong},
  journal={arXiv preprint arXiv:2511.19418},
  year={2025}
}

@inproceedings{liu2025visual,
  title={Visual-rft: Visual reinforcement fine-tuning},
  author={Liu, Ziyu and Sun, Zeyi and Zang, Yuhang and Dong, Xiaoyi and Cao, Yuhang and Duan, Haodong and Lin, Dahua and Wang, Jiaqi},
  booktitle={Proceedings of the IEEE/CVF International Conference on Computer Vision},
  pages={2034--2044},
  year={2025}
}

@article{shen2025vlm,
  title={Vlm-r1: A stable and generalizable r1-style large vision-language model},
  author={Shen, Haozhan and Liu, Peng and Li, Jingcheng and Fang, Chunxin and Ma, Yibo and Liao, Jiajia and Shen, Qiaoli and Zhang, Zilun and Zhao, Kangjia and Zhang, Qianqian and others},
  journal={arXiv preprint arXiv:2504.07615},
  year={2025}
}

@article{huang2025vision,
  title={Vision-r1: Incentivizing reasoning capability in multimodal large language models},
  author={Huang, Wenxuan and Jia, Bohan and Zhai, Zijie and Cao, Shaosheng and Ye, Zheyu and Zhao, Fei and Xu, Zhe and Tang, Xu and Hu, Yao and Lin, Shaohui},
  journal={arXiv preprint arXiv:2503.06749},
  year={2025}
}

@article{chen2025mint,
  title={Mint-cot: Enabling interleaved visual tokens in mathematical chain-of-thought reasoning},
  author={Chen, Xinyan and Zhang, Renrui and Jiang, Dongzhi and Zhou, Aojun and Yan, Shilin and Lin, Weifeng and Li, Hongsheng},
  journal={arXiv preprint arXiv:2506.05331},
  year={2025}
}

@inproceedings{lei2025scaffolding,
  title={Scaffolding coordinates to promote vision-language coordination in large multi-modal models},
  author={Lei, Xuanyu and Yang, Zonghan and Chen, Xinrui and Li, Peng and Liu, Yang},
  booktitle={Proceedings of the 31st International Conference on Computational Linguistics},
  pages={2886--2903},
  year={2025}
}

@article{yu2025introducing,
  title={Introducing visual perception token into multimodal large language model},
  author={Yu, Runpeng and Ma, Xinyin and Wang, Xinchao},
  journal={arXiv preprint arXiv:2502.17425},
  year={2025}
}

@inproceedings{bigverdi2025perception,
  title={Perception tokens enhance visual reasoning in multimodal language models},
  author={Bigverdi, Mahtab and Luo, Zelun and Hsieh, Cheng-Yu and Shen, Ethan and Chen, Dongping and Shapiro, Linda G and Krishna, Ranjay},
  booktitle={Proceedings of the Computer Vision and Pattern Recognition Conference},
  pages={3836--3845},
  year={2025}
}

@article{cub,
  title={The caltech-ucsd birds-200-2011 dataset},
  author={Wah, Catherine and Branson, Steve and Welinder, Peter and Perona, Pietro and Belongie, Serge},
  journal={Technical Report CNS-TR-2011-001},
  year={2011},
}

@inproceedings{gqa,
  title={Gqa: A new dataset for real-world visual reasoning and compositional question answering},
  author={Hudson, Drew A and Manning, Christopher D},
  booktitle={Proceedings of the IEEE/CVF conference on computer vision and pattern recognition},
  pages={6700--6709},
  year={2019},
}

@article{openimages,
  title={The open images dataset v4: Unified image classification, object detection, and visual relationship detection at scale},
  author={Kuznetsova, Alina and Rom, Hassan and Alldrin, Neil and Uijlings, Jasper and Krasin, Ivan and Pont-Tuset, Jordi and Kamali, Shahab and Popov, Stefan and Malloci, Matteo and Kolesnikov, Alexander and others},
  journal={International journal of computer vision},
  volume={128},
  number={7},
  pages={1956--1981},
  year={2020},
  publisher={Springer},
}

@inproceedings{sroie,
  title={Icdar2019 competition on scanned receipt ocr and information extraction},
  author={Huang, Zheng and Chen, Kai and He, Jianhua and Bai, Xiang and Karatzas, Dimosthenis and Lu, Shijian and Jawahar, CV},
  booktitle={2019 International Conference on Document Analysis and Recognition (ICDAR)},
  pages={1516--1520},
  year={2019},
  organization={IEEE},
}

@article{vsr,
  title={Visual spatial reasoning},
  author={Liu, Fangyu and Emerson, Guy and Collier, Nigel},
  journal={Transactions of the Association for Computational Linguistics},
  volume={11},
  pages={635--651},
  year={2023},
  publisher={MIT Press One Broadway, 12th Floor, Cambridge, Massachusetts 02142, USA~…}
}

@misc{msvqa,
      title={Multimodal Continual Learning with MLLMs from Multi-scenario Perspectives}, 
      author={Kai Jiang and Siqi Huang and Xiangyu Chen and Jiawei Shao and Hongyuan Zhang and Ping Luo and Xuelong Li},
      year={2026},
      eprint={2511.18507},
      archivePrefix={arXiv},
      primaryClass={cs.CV},
      url={https://arxiv.org/abs/2511.18507}, 
}

@inproceedings{coco2017,
  title={Microsoft coco: Common objects in context},
  author={Lin, Tsung-Yi and Maire, Michael and Belongie, Serge and Hays, James and Perona, Pietro and Ramanan, Deva and Doll{\'a}r, Piotr and Zitnick, C Lawrence},
  booktitle={European conference on computer vision},
  pages={740--755},
  year={2014},
  organization={Springer},
}

@inproceedings{objects365,
  title={Objects365: A large-scale, high-quality dataset for object detection},
  author={Shao, Shuai and Li, Zeming and Zhang, Tianyuan and Peng, Chao and Yu, Gang and Zhang, Xiangyu and Li, Jing and Sun, Jian},
  booktitle={Proceedings of the IEEE/CVF international conference on computer vision},
  pages={8430--8439},
  year={2019},
}

@article{ruod,
title = {Rethinking general underwater object detection: Datasets, challenges, and solutions},
journal = {Neurocomputing},
volume = {517},
pages = {243-256},
year = {2023},
issn = {0925-2312},
doi = {https://doi.org/10.1016/j.neucom.2022.10.039},
author = {Chenping Fu and Risheng Liu and Xin Fan and Puyang Chen and Hao Fu and Wanqi Yuan and Ming Zhu and Zhongxuan Luo},
}

@article{visdrone,
  title={Detection and tracking meet drones challenge},
  author={Zhu, Pengfei and Wen, Longyin and Du, Dawei and Bian, Xiao and Fan, Heng and Hu, Qinghua and Ling, Haibin},
  journal={IEEE Transactions on Pattern Analysis and Machine Intelligence},
  volume={44},
  number={11},
  pages={7380--7399},
  year={2021},
  publisher={IEEE},
}

@misc{qwen3_5,
    title  = {{Qwen3.5}: Towards Native Multimodal Agents},
    author = {{Qwen Team}},
    year   = {2026},
    month  = {February},
    url    = {https://qwen.ai/blog?id=qwen3.5}
}

@inproceedings{RN,
  title={Rectified Noise: A Generative Model Using Positive-incentive Noise},
  author={Gu, Zhenyu and Xu, Yanchen and Huang, Sida and Guo, Yubin and Zhang, Hongyuan},
  booktitle={Proceedings of the AAAI Conference on Artificial Intelligence},
  volume={40},
  pages={4357--4365},
  year={2026}
}

@inproceedings{MuNG,
  title={Explore how to inject beneficial noise in mllms},
  author={Zhu, Ruishu and Huang, Sida and Jiao, Ziheng and Zhang, Hongyuan},
  booktitle={Proceedings of the AAAI Conference on Artificial Intelligence},
  volume={40},
  pages={29150--29158},
  year={2026}
}

@inproceedings{ViewMask,
  title={ViewMask-1-to-3: Multi-View Consistent Image Generation via Multimodal Discrete Diffusion Models},
  author={Zhu, Ruishu and Huang, Zhihao and Sun, Jiacheng and Luo, Ping and Zhang, Hongyuan and Li, Xuelong},
  booktitle={Proceedings of the 43rd International Conference on Machine Learning (ICML)},
  year={2026}
}

@inproceedings{MiN,
  title={Mixture of Noise for Pre-Trained Model-Based Class-Incremental Learning}, 
  author={Kai Jiang and Zhengyan Shi and Dell Zhang and Hongyuan Zhang and Xuelong Li},
  booktitle={Advances in neural information processing systems},
  volume={38},
  pages={44776--44802},
  year={2026},
}

@article{VPN,
    title={Variational Positive-incentive Noise: How Noise Benefits Models}, 
    author={Hongyuan Zhang and Sida Huang and Yubin Guo and Xuelong Li},
    year={2025},
    journal={IEEE Transactions on Pattern Analysis and Machine Intelligence},
}
\bibliographystyle{iclr2026_conference}

\appendix
\section*{Appendix}

\numberwithin{equation}{section}
\numberwithin{table}{section}
\numberwithin{figure}{section}

\section{Dataset and Evaluation Details}
\label{appdatasets}

\textbf{The evaluation suite is designed to test whether \textsc{Lens} can select fine-grained visual evidence across answer generation and spatial localization.}
Table~\ref{tabappdatasets} summarizes the datasets, task forms, split sizes, input formats, and evaluation metrics. The six VQA datasets cover attribute recognition, compositional reasoning, OCR, spatial relation understanding, and multi-choice activity understanding. The four grounding datasets cover common objects, long-tail objects, underwater objects, and dense aerial scenes. This combination is important because visual evidence purification should improve both the answer produced from selected evidence and the localization of that evidence.

For VQA evaluation, each prediction is compared with the reference answer by Qwen3.5-Flash. The judge receives the question, the reference answer, and the model prediction, then returns a scalar score in $[0,1]$. We average the sample scores and report the normalized percentage score. This protocol gives partial credit when a prediction is semantically close to the reference, which is useful for open-ended VQA and OCR-style answers.

For grounding evaluation, we first extract all predicted boxes from the model output. Each predicted category is mapped to the closest category in the dataset vocabulary by Qwen3.5-Flash. Predictions without a valid category match are removed. We then match predicted boxes and ground-truth boxes of the same mapped category at IoU threshold 0.5 and compute the dataset-level F1 score. This evaluation penalizes both missed objects and unsupported detections, so it directly reflects whether the model localizes the complete evidence set.

\textbf{The two benchmark groups provide complementary evidence for the central claim.}
VQA measures whether selected visual evidence supports answer generation, while grounding measures whether the selected evidence is spatially correct. Consistent gains on both groups therefore indicate that \textsc{Lens} improves question-conditioned visual selectivity rather than only adapting to one output format.

\begin{table*}[htbp]
\renewcommand{\arraystretch}{1.08}
\centering
\small
\setlength{\tabcolsep}{5.0pt}
\caption{Dataset and evaluation details. Train and validation sizes are counted from our annotation files. Judge score denotes VQA scoring by Qwen3.5-Flash, and F1@0.5 denotes class-normalized grounding F1.}
\label{tabappdatasets}
\begin{tabularx}{\linewidth}{@{}llrrXX@{}}
\toprule
\textbf{Dataset} & \textbf{Task} & \textbf{Train} & \textbf{Val} & \textbf{Input} & \textbf{Metric} \\
\midrule
\multicolumn{6}{@{}l}{\textbf{VQA}}\\
CUB & Attribute & 10,056 & 492 & Image question & Judge score \\
GQA & Compositional & 98,149 & 978 & Image question & Judge score \\
OpenImages & Open domain & 43,053 & 945 & Image question & Judge score \\
SROIE & OCR & 2,486 & 686 & Document query & Judge score \\
VSR & Spatial & 3,376 & 404 & Image statement & Judge score \\
MSVQA & Multi choice & 13,603 & 500 & Image question & Judge score \\
\midrule
\multicolumn{6}{@{}l}{\textbf{Grounding}}\\
COCO2017 & Common objects & 118,287 & 1,000 & Localization prompt & F1@0.5 \\
Objects365 & Long tail objects & 200,000 & 1,000 & Localization prompt & F1@0.5 \\
RUOD & Underwater objects & 13,744 & 249 & Localization prompt & F1@0.5 \\
VisDrone & Aerial objects & 6,004 & 249 & Localization prompt & F1@0.5 \\
\bottomrule
\end{tabularx}
\end{table*}

\section{Implementation Details}
\label{appimplementation}

\textbf{The implementation keeps the MLLM backbone and the visual token sequence interface unchanged.}
Unless otherwise stated, experiments use Qwen3-VL-4B as the base model and train on 8 NVIDIA H100 80G GPUs. Input images are resized under a maximum pixel budget of $1000000$ and a minimum pixel budget of $3136$. Spatial sizes are aligned to the visual patch stride of 32. Thus the number of visual tokens is bounded by the resized patch grid and remains below about 1000 tokens, while very small images still keep at least three visual tokens.

The temporary $\langle mask\rangle$ token is appended to the end of the text input in the probing pass, after the image tokens and question tokens have been formed as in Eq.~\ref{eq:2}. The hidden state at this temporary position is used only to predict the LET scores. The token is removed before answer decoding, so it does not become a persistent reasoning token and does not alter the generation interface.

The probe head is a lightweight one-layer MLP implemented as a linear projection followed by a sigmoid activation. Given the hidden state $\mathbf{h}_{m}$ at the temporary token, the head produces $N$ logits and converts them into the visual evidence prior $\mathbf{a}\in[0,1]^N$. Here $N$ is the number of visual tokens for the current image. The head therefore adds only a small number of parameters and its output is directly aligned with the visual token grid.

The noise generator is a two-branch MLP that predicts feature-aware perturbation parameters for each visual token. Each branch contains a linear layer, a ReLU activation, and a linear output layer. One branch predicts $\boldsymbol{\mu}_i$ and the other predicts the raw scale parameter for $\boldsymbol{\sigma}_i$. The final latent perturbation is sampled by reparameterization,
\begin{align}
\mathbf{r}_i
=
\boldsymbol{\mu}_i
+
\boldsymbol{\sigma}_i\odot\boldsymbol{\epsilon}_i,
\quad
\boldsymbol{\epsilon}_i\sim\mathcal{N}(\mathbf{0},\mathbf{I}).
\end{align}
Latent suppression is applied after the visual encoder has produced visual token embeddings and before these embeddings are merged with the text tokens for decoding. This position lets \textsc{Lens} suppress visual distractors without changing image preprocessing, tokenizer behavior, language decoding, or the output format.

We train for 1 epoch with batch size 128, learning rate $4\times10^{-5}$, AdamW optimizer, warmup ratio 0.03, weight decay 0, and bf16 mixed precision. The loss weight for evidence supervision is $\beta=0.2$, and the deterministic inference threshold is $\tau=0.5$. The same preprocessing and decoding settings are used across baselines unless a baseline requires its own official setting.

\textbf{These details make the extra computation localized and reproducible.}
The only additional forward computation is the short probing pass used to estimate $\mathbf{a}$. After that, the model performs standard autoregressive decoding with the same token sequence length, using purified visual embeddings $\widetilde{\mathbf{V}}$ instead of the original visual embeddings $\mathbf{V}$.

\section{Evidence Supervision Construction}
\label{appevidence}

\textbf{\textsc{Lens} uses object-level boxes to supervise the question-conditioned \emph{Lens Evidence Token} without requiring chain-of-thought rationales or pixel-level masks.}
Each training sample provides an image, a question or localization prompt, an answer, and a set of question-relevant boxes. We first resize the image with the same preprocessing used by the MLLM. If the original image has width $W$ and height $H$, and the resized image has width $\widetilde{W}$ and height $\widetilde{H}$, every box $B=(x_1,y_1,x_2,y_2)$ is mapped to
\begin{align}
\widetilde{B}
=
\left(
\left\lfloor x_1\frac{\widetilde{W}}{W}\right\rfloor,
\left\lfloor y_1\frac{\widetilde{H}}{H}\right\rfloor,
\left\lfloor x_2\frac{\widetilde{W}}{W}\right\rfloor,
\left\lfloor y_2\frac{\widetilde{H}}{H}\right\rfloor
\right).
\end{align}
The resized image is divided into a 32 by 32 patch grid. Let patch $\Omega_i$ be the spatial area of visual token $\mathbf{v}_i$. The binary evidence label is
\begin{align}
z_i
=
\mathbb{I}
\left[
\max_{\widetilde{B}\in\mathcal{B}_{Q}}
\mathrm{area}(\Omega_i\cap\widetilde{B})
>
0
\right].
\end{align}
The labels are flattened from left to right and top to bottom, which gives a token-level target $\mathbf{z}\in\{0,1\}^{N}$ aligned with the visual token order.

Multiple question-relevant objects are handled by taking the union over boxes. If boxes overlap, a patch is still labeled once, so overlapping annotations do not overweight a region. Small objects are retained as long as they intersect at least one patch. This rule is important for CUB attributes, OCR fields, VisDrone objects, and RUOD objects, where the evidence can occupy a small part of the image.

For OCR-oriented samples such as SROIE, the evidence boxes correspond to text regions needed to answer the question. Patches intersecting those text boxes are labeled as positive, while unrelated document regions remain negative. For common object grounding datasets such as COCO2017, Objects365, RUOD, and VisDrone, the prompt asks the model to locate common or dataset-defined objects, so all annotated target boxes in the sample are treated as the relevant evidence set.

We remove invalid boxes with non-positive width or height after resizing. If a sample has no valid box or produces no positive patch after preprocessing, its LET loss is masked out for that sample. The answer loss can still be used when the answer annotation is valid. This filtering avoids training the probe with empty or contradictory evidence labels.

\counterwithout{figure}{section}
\setcounter{figure}{0}
\renewcommand{\thefigure}{H.\arabic{figure}}
\providecommand{\theHfigure}{\thefigure}
\renewcommand{\theHfigure}{H.\arabic{figure}}

\begin{figure*}[thbp]
    \centering
    \includegraphics[width=.96\linewidth,height=.62\textheight,keepaspectratio]{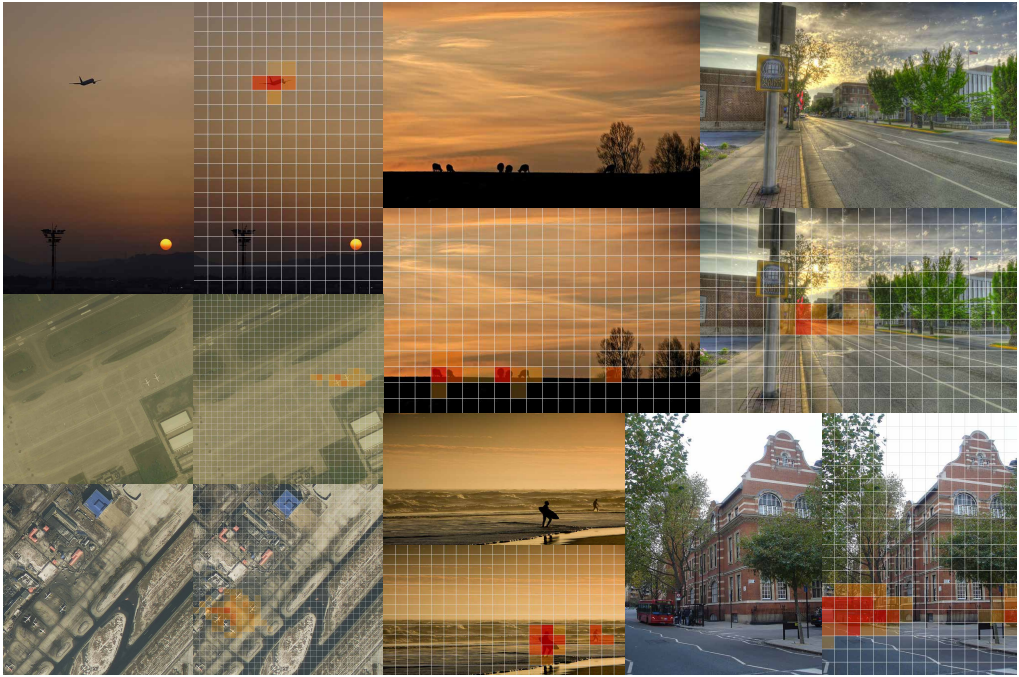}
    \caption{Additional small object visualizations. The predicted LET scores focus on small targets such as distant aircraft, far people, road signs, small aerial objects, pedestrians, and vehicles.}
    \label{figappsmall}
\end{figure*}

\textbf{This construction turns inexpensive box supervision into token-level LET supervision.}
It is weaker than dense segmentation, but it is sufficient for the goal of \textsc{Lens}, since latent suppression only needs to distinguish likely evidence tokens from broad distractors. The supervision also remains question-conditioned because the positive boxes are selected according to the current question or localization prompt rather than image saliency alone.

\section{Reinforcement Fine-Tuning Details}
\label{apprft}

\textbf{Reinforcement fine-tuning refines the LET scores as a set-level policy rather than adding a new reasoning module.}
After supervised training, each LET score $a_i$ is interpreted as the Bernoulli parameter for token selection,
\begin{align}
\pi_{\theta}(\mathbf{b}\mid I,Q)
=
\prod_{i=1}^{N}
a_i^{b_i}
(1-a_i)^{1-b_i},
\quad
b_i\in\{0,1\}.
\end{align}
During training, $\mathbf{b}$ is sampled stochastically. This sampling exposes the model to boundary tokens whose LET scores are uncertain, which acts as data augmentation for evidence selection. During inference, we do not sample. We instead use the deterministic threshold $\tau=0.5$ and preserve tokens with $a_i\geq\tau$.

\begin{figure*}[tbp]
    \centering
    \includegraphics[width=\linewidth]{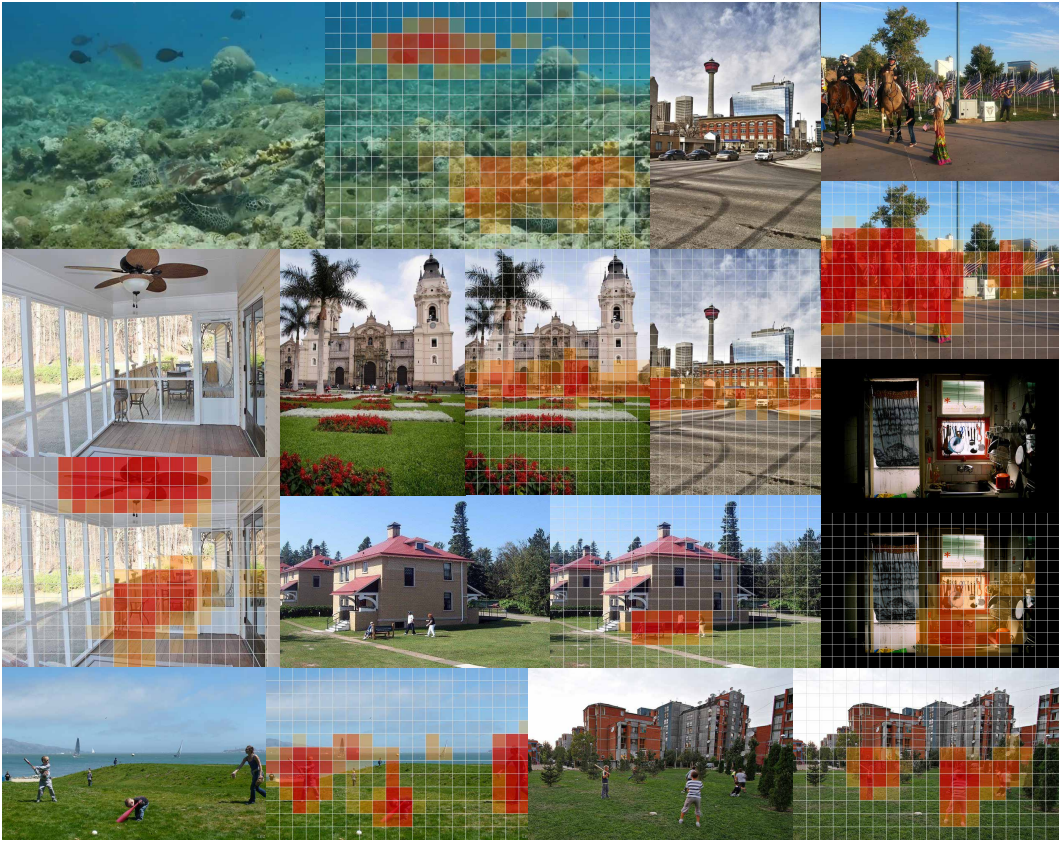}
    \caption{Additional visualizations for occluded, hidden, and dense scenes. The examples include underwater targets, cluttered street scenes, indoor occlusion, building scenes, and crowded outdoor scenes.}
    \label{figappclutter}
\end{figure*}

We optimize the evidence policy with GRPO. For each image-question pair, we sample $K=4$ evidence masks from the current Bernoulli policy. The evidence reward follows the set F1 score between the sampled mask and the target label,
\begin{align}
R_{\mathrm{ev}}(\mathbf{b}^{(k)},\mathbf{z})
=
\frac{
2\sum_{i=1}^{N}b_i^{(k)}z_i
}{
\sum_{i=1}^{N}b_i^{(k)}
+
\sum_{i=1}^{N}z_i
+
\varepsilon
}.
\end{align}
The group baseline is the mean reward of the $K$ samples, and the normalized advantage is
\begin{align}
\widehat{A}^{(k)}
=
\frac{
R_{\mathrm{ev}}(\mathbf{b}^{(k)},\mathbf{z})
-
\frac{1}{K}\sum_{\ell=1}^{K}R_{\mathrm{ev}}(\mathbf{b}^{(\ell)},\mathbf{z})
}{
\mathrm{std}_{\ell}\left(R_{\mathrm{ev}}(\mathbf{b}^{(\ell)},\mathbf{z})\right)
+
\varepsilon
}.
\end{align}
The policy objective is
\begin{align}
\mathcal{L}_{\mathrm{GRPO}}
=
-
\frac{1}{K}
\sum_{k=1}^{K}
\widehat{A}^{(k)}
\log
\pi_{\theta}(\mathbf{b}^{(k)}\mid I,Q)
+
\lambda
D_{\mathrm{KL}}
\left(
\pi_{\theta}(\cdot\mid I,Q)
\|
\pi_{\mathrm{SFT}}(\cdot\mid I,Q)
\right),
\end{align}
where $\lambda=0.01$. The KL term keeps the policy close to the supervised LET policy and prevents reward optimization from collapsing to overly sparse or overly dense masks.

For answer-level reinforcement baselines, we use the same task reward implementation for all RL methods. The reward dispatches by dataset type. General VQA uses normalized text or choice accuracy, SROIE uses OCR text similarity, MSVQA uses yes-no matching, count accuracy, list F1, or choice accuracy, and grounding uses box F1 at IoU 0.5 after category normalization. This shared reward makes the RL comparison fair, while the additional evidence reward above is specific to refining \textsc{Lens} as a token selection policy.

The sampled mask is used only during training. Given $\mathbf{b}^{(k)}$, selected tokens are preserved and unselected tokens receive stronger latent suppression. At inference, the deterministic mask from $\tau$ is used to construct $\widetilde{\mathbf{V}}$. This difference avoids test-time randomness while preserving the training benefit of stochastic exploration around uncertain evidence boundaries.

\textbf{The RFT stage therefore improves the completeness of the selected evidence set.}
Supervised learning labels each token independently, whereas GRPO rewards the whole selected set. This is better aligned with VQA and grounding, where the model often needs several supporting patches or multiple object instances to answer correctly.

\section{Full Baseline and Reproduction Protocol}
\label{appbaseline}

\textbf{All baselines are reproduced under the same backbone, data, and evaluation protocol whenever the official implementation permits it.}
For each trainable method, we replace the original backbone with Qwen3-VL-4B and use the same training split described in Appendix~\ref{appdatasets}. We keep the default hyperparameters in the released code unless the backbone replacement requires a dimension or tokenizer adjustment. This protocol isolates the method design from changes in data scale, model capacity, and evaluation scripts.

\paragraph{Vanilla}
\textbf{Vanilla measures the zero-training capability of the Qwen3-VL-4B backbone.}
We directly run inference with the shared preprocessing, prompting, decoding, and evaluation scripts. This baseline shows how much visual understanding and grounding ability is already present before any task-specific adaptation.

\paragraph{SFT}
\textbf{SFT measures the effect of full-parameter supervised adaptation under the same data budget as \textsc{Lens}.}
We fine-tune Qwen3-VL-4B on the same training data and use the implementation settings in Appendix~\ref{appimplementation}. This baseline separates the gain from ordinary supervised training from the gain brought by question-conditioned evidence probing and latent suppression.

\paragraph{Visual-RFT}
\textbf{Visual-RFT is reproduced as a reinforcement fine-tuning baseline with visual verifiable rewards.}
We use the official Visual-RFT implementation~\citep{liu2025visual}, replace the base MLLM with Qwen3-VL-4B, and keep its default GRPO hyperparameters. The method samples multiple responses with reasoning tokens and final answers, then updates the policy with task rewards such as visual classification correctness or localization quality.

\paragraph{VLM-R1}
\textbf{VLM-R1 evaluates an R1-style reinforcement learning pipeline for vision-language tasks.}
We use the official VLM-R1 implementation~\citep{shen2025vlm} with Qwen3-VL-4B and the same training data. Its policy update follows the released GRPO setting and optimizes visual understanding or grounding outputs through verifiable rewards.

\paragraph{PAPO}
\textbf{PAPO tests whether perception-aware policy optimization improves multimodal reasoning under the same data setting.}
We use the official PAPO implementation~\citep{wang2025perception}, replace the backbone with Qwen3-VL-4B, and keep the default PAPO hyperparameters. The method adds an implicit perception loss to RLVR by comparing rollout likelihoods under clean and corrupted visual inputs, encouraging the policy to rely on informative visual evidence while reasoning.

\paragraph{VPT}
\textbf{VPT is reproduced without external visual encoders to keep the comparison fair.}
We use the official VPT implementation~\citep{yu2025introducing} with Qwen3-VL-4B and the same training data. VPT introduces visual perception tokens that can trigger extra region perception or visual re-encoding. In our reproduction, we only use the CLIP style mode with the model native visual encoder and do not add DINO, SAM, or other external encoders, because those encoders would provide additional visual knowledge unavailable to the other baselines.

\paragraph{LVR}
\textbf{LVR is kept as a two-stage latent visual reasoning baseline.}
We use the official LVR implementation, replace the base MLLM with Qwen3-VL-4B, and preserve both stages in the original recipe. The first stage performs supervised fine-tuning that jointly learns latent visual reasoning and text generation, while the second stage applies reinforcement learning to refine the latent reasoning process with response-level rewards.

\paragraph{DMLR}
\textbf{DMLR is evaluated as a training-free test-time latent reasoning method.}
Since DMLR~\citep{liu2025reasoning} does not require additional training, we apply the official inference procedure on the SFT checkpoint. It refines latent think tokens at test time through confidence-guided latent updates and dynamically injects selected visual features into the latent reasoning process.

\paragraph{VisMem}
\textbf{VisMem is reproduced with its two-stage latent memory training pipeline.}
We use the official VisMem implementation~\citep{yu2025vismem}, replace the base MLLM with Qwen3-VL-4B, and keep the original memory learning procedure. The method first learns short-term and long-term latent vision memories, then learns how to invoke these memories during inference to support perceptual retention and semantic consolidation.

\textbf{This reproduction design makes the comparison conservative for \textsc{Lens}.}
All trainable baselines use the same data and base model size, while method-specific training stages are retained when they are part of the official recipe. As a result, the comparison mainly reflects whether a method adds textual reasoning, visual prompts, latent reasoning, latent memories, test-time latent updates, or question-conditioned visual evidence suppression. \textsc{Lens} keeps the backbone interface unchanged and learns to predict and suppress question-irrelevant visual tokens.

\begin{figure*}[tbp]
    \centering
    \includegraphics[width=\linewidth]{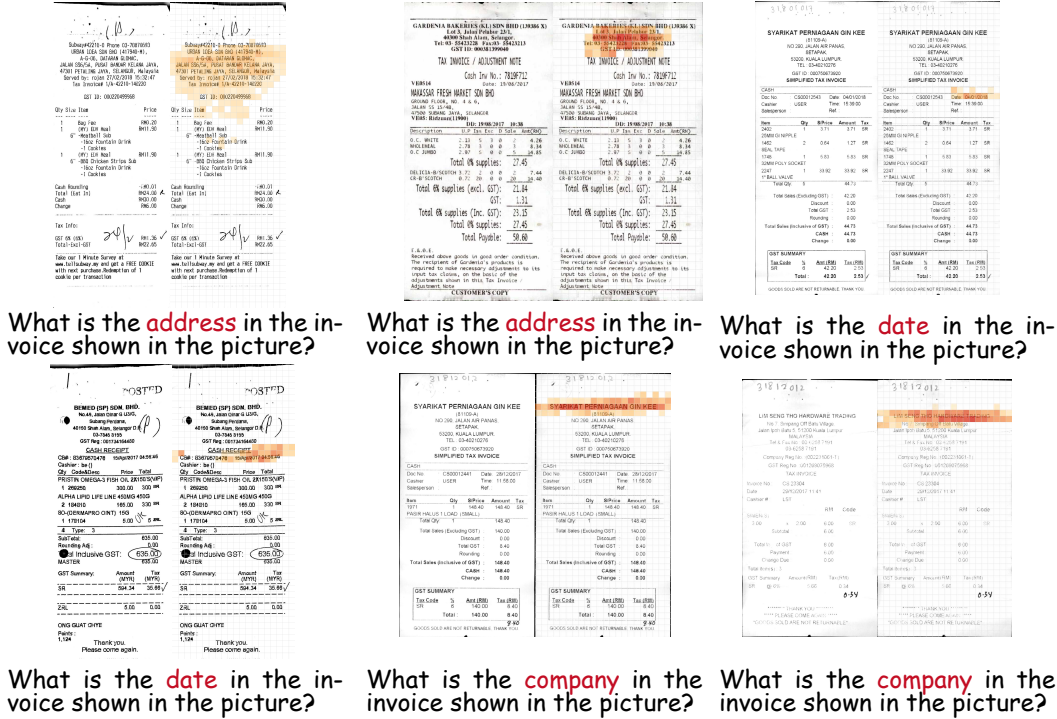}
    \caption{Additional OCR visualizations. The predicted LET scores move to the text field required by the question, including address, date, and company information.}
    \label{figappocr}
\end{figure*}

\section{Additional Ablation Studies}
\label{appadditionalablation}

\textbf{The additional ablations show that \textsc{Lens} is not driven by a fragile hyperparameter choice.}
We vary the inference threshold $\tau$ in $\{0.3,0.5,0.7\}$ and the LET loss weight $\beta$ in $\{0.1,0.2,0.4\}$. The backbone, training data, decoding setting, and evaluation scripts are kept unchanged. Table~\ref{tabappparamablation} reports all benchmark scores together with VQA, grounding, and overall averages.

\begin{table*}[tbp]
\renewcommand{\arraystretch}{1.08}
\centering
\small
\setlength{\tabcolsep}{3.0pt}
\caption{Additional ablations on the evidence threshold $\tau$ and the LET loss weight $\beta$. VQA Avg. averages the six VQA datasets, Grounding Avg. averages the four grounding datasets, and Avg. averages all ten datasets.}
\label{tabappparamablation}
\resizebox{\linewidth}{!}{
\begin{tabular}{lccccccccccccc}
\toprule
\textbf{Setting} & CUB & GQA & OpenImg & SROIE & VSR & MSVQA & \textbf{VQA Avg.} & COCO & Obj365 & RUOD & Visdrone & \textbf{Grounding Avg.} & \textbf{Avg.} \\
\midrule
\multicolumn{14}{@{}l}{\textbf{Threshold $\tau$}}\\
$\tau=0.3$ & 88.87 & 78.45 & 85.84 & 93.16 & 77.23 & 68.10 & 81.94 & 53.20 & 43.10 & 69.57 & 46.29 & 53.04 & 70.38\\
$\tau=0.5$ & 91.50 & 79.01 & 86.89 & 95.20 & 80.69 & 66.39 & \textbf{83.28} & 52.00 & 44.10 & 70.05 & 44.09 & 52.56 & 70.99\\
$\tau=0.7$ & 90.69 & 78.75 & 87.05 & 92.73 & 78.47 & 70.54 & 83.04 & 51.79 & 44.49 & 71.34 & 45.02 & \textbf{53.16} & \textbf{71.09}\\
\midrule
\multicolumn{14}{@{}l}{\textbf{LET loss weight $\beta$}}\\
$\beta=0.1$ & 89.75 & 78.62 & 85.86 & 94.05 & 78.65 & 65.89 & 82.14 & 50.77 & 43.65 & 69.68 & 44.27 & 52.09 & 70.12\\
$\beta=0.2$ & 91.50 & 79.01 & 86.89 & 95.20 & 80.69 & 66.39 & \textbf{83.28} & 52.00 & 44.10 & 70.05 & 44.09 & 52.56 & \textbf{70.99}\\
$\beta=0.4$ & 90.59 & 77.84 & 84.95 & 92.15 & 77.45 & 67.18 & 81.69 & 53.88 & 44.87 & 71.36 & 44.92 & \textbf{53.76} & 70.52\\
\bottomrule
\end{tabular}
}
\end{table*}

\textbf{The threshold $\tau$ mainly controls the balance between preserving weak evidence and removing distractors.}
At inference, tokens with $a_i\geq\tau$ are preserved and tokens with lower LET scores receive latent suppression. The preserved token ratio and the mean suppression strength are
\begin{align}
\rho_{\tau}
&=
\frac{1}{N}
\sum_{i=1}^{N}
\mathbb{I}\left[a_i\geq\tau\right],
\\
\bar{g}_{\tau}
&=
\frac{1}{N}
\sum_{i=1}^{N}
\frac{1}{\tau}
\mathrm{ReLU}\left(\tau-a_i\right).
\end{align}
When $\tau=0.3$, the intervention is conservative and keeps more marginal patches active. This protects possible evidence, but it also leaves distractors in the context, giving an overall average of 70.38. Increasing $\tau$ to 0.5 raises the VQA average to 83.28 and gives the best results on CUB, GQA, SROIE, and VSR. This shows that moderate suppression helps fine-grained answer generation. With $\tau=0.7$, the grounding average reaches 53.16 and the overall average reaches 71.09, but SROIE and VSR drop compared with $\tau=0.5$. This pattern indicates that aggressive suppression can help localization while perturbing small OCR or spatial cues.

\textbf{The loss weight $\beta$ shows a similar tradeoff between LET supervision and answer flexibility.}
The supervised objective is
\begin{align}
\mathcal{L}_{\mathrm{SFT}}(\beta)
&=
\mathcal{L}_{\mathrm{ans}}
+
\beta\mathcal{L}_{\mathrm{LET}},
\\
\nabla\mathcal{L}_{\mathrm{SFT}}(\beta)
&=
\nabla\mathcal{L}_{\mathrm{ans}}
+
\beta\nabla\mathcal{L}_{\mathrm{LET}}.
\end{align}
When $\beta=0.1$, the LET score vector is weakly supervised and the overall average is 70.12. The default $\beta=0.2$ obtains the best overall average of 70.99 and the best VQA average of 83.28, which suggests that the LET scores are strong enough to guide suppression without dominating answer learning. Increasing $\beta$ to 0.4 improves the grounding average to 53.76, including the best scores on COCO, Object365, RUOD, and VisDrone. However, the VQA average decreases to 81.69. This means stronger box driven supervision improves spatial selectivity, but can reduce flexibility for open ended answer generation.

\textbf{These results support the default setting used in the main experiments.}
The best overall threshold is only 0.10 points above $\tau=0.5$, while $\tau=0.5$ gives the strongest VQA average and a balanced grounding score. For $\beta$, the default value gives the best overall score and avoids the VQA degradation observed at $\beta=0.4$. Therefore, the main results are not caused by a narrow parameter optimum. They reflect the intended mechanism of \textsc{Lens}, question-conditioned LET scores preserve answer-supporting patches and latent suppression weakens visually redundant regions.

\section{Computational Cost Analysis}
\label{appcostanalysis}

\textbf{\textsc{Lens} adds a localized probing cost while keeping the decoding interface unchanged.}
Let $N$ be the number of visual tokens, $M$ be the number of text tokens, $L$ be the output length, and $d$ be the hidden dimension. A standard MLLM mainly pays the autoregressive decoding cost $C_{\mathrm{decode}}(N,M,L)$. \textsc{Lens} adds one evidence probing pass and one token-wise latent suppression operation,
\begin{align}
C_{\textsc{Lens}}
\approx
C_{\mathrm{probe}}(N,M)
+
C_{\mathrm{sup}}(N,d)
+
C_{\mathrm{decode}}(N,M,L).
\end{align}
The purified visual sequence keeps the same length as the original visual sequence,
\begin{align}
\left|\widetilde{\mathbf{V}}\right|
=
\left|\mathbf{V}\right|
=
N.
\end{align}
Thus, \textsc{Lens} does not add visual tokens to the decoder and does not require an external detector, an image generator, or test-time latent optimization.

\begin{table*}[tbp]
\renewcommand{\arraystretch}{1.10}
\centering
\small
\setlength{\tabcolsep}{7.0pt}
\caption{Inference cost comparison on the VQA test split. TTFT denotes Time To First Token, TPOT denotes Time Per Output Token, and Avg. Length denotes the generated answer length. Timing is measured on one H100 GPU with five warm-up rounds and ten measured rounds.}
\label{tabappcost}
\resizebox{\linewidth}{!}{
\begin{tabular}{lccccccc}
\toprule
\multirow{2}{*}{\textbf{Method}} &
\multicolumn{3}{c}{\textbf{TTFT $\mathrm{ms}$}} &
\multicolumn{3}{c}{\textbf{TPOT $\mathrm{ms}$}} &
\multirow{2}{*}{\textbf{Avg. Length}} \\
\cmidrule(lr){2-4}\cmidrule(lr){5-7}
& \textbf{bs=1} & \textbf{bs=8} & \textbf{bs=16} & \textbf{bs=1} & \textbf{bs=8} & \textbf{bs=16} & \\
\midrule
GT & -- & -- & -- & -- & -- & -- & 14.46 \\
SFT & 71.484$\pm$6.722 & 41.730$\pm$0.688 & 40.840$\pm$0.583 & 35.010$\pm$3.057 & 6.131$\pm$0.398 & 3.574$\pm$0.217 & 14.29\\
Visual-RFT & 188.409$\pm$7.687 & 188.077$\pm$2.807 & 198.696$\pm$2.680 & 35.206$\pm$2.985 & 7.018$\pm$0.617 & 4.537$\pm$0.424 & 20.89 \\
VLM-R1 & 72.165$\pm$6.655 & 41.863$\pm$0.501 & 39.271$\pm$0.620 & 34.015$\pm$2.039 & 6.258$\pm$0.895 & 4.755$\pm$0.260 & 17.08 \\
LVR & 88.497$\pm$4.099 & 74.970$\pm$0.749 & 77.040$\pm$0.367 & 31.937$\pm$1.785 & 6.512$\pm$0.269 & 4.089$\pm$0.144 & 28.44 \\
PAPO & 73.218$\pm$7.097 & 41.087$\pm$0.735 & 39.268$\pm$0.599 & 35.417$\pm$3.625 & 6.894$\pm$0.735 & 4.659$\pm$0.975 & 14.91 \\
VPT & 172.081$\pm$22.503 & 128.372$\pm$7.544 & 119.615$\pm$3.237 & 105.011$\pm$6.952 & 29.409$\pm$1.635 & 17.312$\pm$0.791 & 34.88 \\
VisMem & 151.129$\pm$24.425 & 97.919$\pm$7.532 & 89.136$\pm$5.133 & 64.379$\pm$5.983 & 38.455$\pm$3.589 & 23.512$\pm$2.333 & 22.41 \\
\textsc{Lens} & 122.163$\pm$11.163 & 66.886$\pm$0.995 & 65.412$\pm$0.832 & 34.156$\pm$4.239 & 6.161$\pm$0.539 & 3.494$\pm$0.268 & 15.98 \\
\bottomrule
\end{tabular}
}
\end{table*}

\textbf{The timing data show that the extra cost of \textsc{Lens} is paid before decoding rather than during each generated token.}
Using SFT as the closest backbone matched baseline, the TTFT overhead is
\begin{align}
\Delta_{\mathrm{TTFT}}(b)
=
\mathrm{TTFT}_{\textsc{Lens}}(b)
-
\mathrm{TTFT}_{\mathrm{SFT}}(b),
\end{align}
which gives 50.68 ms, 25.16 ms, and 24.57 ms for bs=1, bs=8, and bs=16. In contrast, the per token change is
\begin{align}
\Delta_{\mathrm{TPOT}}(b)
=
\mathrm{TPOT}_{\textsc{Lens}}(b)
-
\mathrm{TPOT}_{\mathrm{SFT}}(b),
\end{align}
which gives -0.85 ms, 0.03 ms, and -0.08 ms for the same batch sizes. The near zero TPOT change confirms that \textsc{Lens} does not make autoregressive decoding slower after the visual evidence has been purified.

\textbf{The comparison with reasoning and memory baselines further supports the efficiency claim.}
Visual-RFT has much larger TTFT and a longer average output length of 20.89. LVR and VisMem also increase answer length because they rely on latent reasoning or memory usage. VPT has the largest TPOT cost, reaching 105.011 ms at bs=1, because additional visual perception changes the decoding workload. By contrast, \textsc{Lens} keeps the average length close to SFT and GT, with 15.98 tokens compared with 14.29 and 14.46. This shows that the gain does not come from longer responses.

\textbf{The cost profile matches the design goal of visual evidence purification.}
The probing pass estimates $\mathbf{a}$ once, and the suppression operation applies
\begin{align}
\widetilde{\mathbf{v}}_i
=
\mathbf{v}_i
+
\frac{1}{\tau}
\mathrm{ReLU}\left(\tau-a_i\right)
\mathbf{r}_i,
\end{align}
before standard decoding. Since this operation is linear in the number of visual tokens and creates no new token sequence, the remaining overhead is predictable. \textsc{Lens} therefore improves visual selectivity with a one-time evidence purification step instead of external tools, test-time latent optimization, or longer reasoning traces.

\section{Additional Qualitative Visualizations}
\label{appadditionalvis}

\textbf{The additional visualizations extend the main qualitative evidence to harder perception settings.}
The main text already shows that \textsc{Lens} can focus on answer-supporting regions in VQA and target objects in grounding. This section adds three groups of cases that are more likely to expose visual redundancy. The purpose is to show that the same question-conditioned LET scores remain useful when the evidence is small, visually ambiguous, or text based.

\begin{itemize}
    \item \textbf{Small or distant targets test whether fine local evidence can be preserved.}
    These cases examine objects that occupy only a few visual patches, where background tokens can easily dilute the useful signal.
    \item \textbf{Occluded, hidden, and dense scenes test whether distractors can be suppressed.}
    These cases examine cluttered images where similar textures, nearby objects, and partial visibility make visual evidence ambiguous.
    \item \textbf{OCR fields test whether the LET scores are truly question-conditioned.}
    These cases examine invoices where the correct region changes with the requested field, e.g., address, date, or company name.
\end{itemize}

\textbf{The small object cases show that \textsc{Lens} can preserve fine evidence even when the target occupies only a few visual patches.}
In Fig.~\ref{figappsmall}, the highlighted regions are concentrated around tiny objects rather than spread over the sky, road, sea, or building background. This behavior is important for the proposed latent suppression mechanism. If the LET scores only followed image saliency, large background areas would remain active and the small target would still be diluted. Instead, the LET assigns higher scores to compact target regions, so latent noise mainly weakens irrelevant context while keeping the local evidence used for the answer.

\textbf{The cluttered scene cases show why visual evidence purification is more suitable than simply adding more reasoning context.}
In Fig.~\ref{figappclutter}, the first underwater example contains a turtle whose texture and color are close to the surrounding reef, making it difficult even for human observers to separate the animal from the background. The learned LET scores still give strong responses around the turtle and other queried regions, while suppressing large areas of water and coral. Similar behavior appears in scenes with riders, flags, building facades, indoor objects, and multiple people. These examples support the central claim that many failures come from distractor interference, not from a lack of visual tokens.

\textbf{The OCR cases demonstrate that the LET scores are conditioned on the question rather than fixed document saliency.}
Invoices contain many visually similar text lines, and a generic document attention pattern may focus on headers, totals, or dense item rows regardless of the question. In Fig.~\ref{figappocr}, the highlighted patches shift according to the requested field. Address questions activate address blocks, date questions activate date lines, and company questions activate merchant names. This behavior matches the objective of \textsc{Lens}, which uses the question to select evidence through the LET scores before latent suppression. As a result, redundant text regions are weakened and the decoder receives a cleaner document representation for OCR style reasoning.

\end{document}